\documentclass[11pt]{article}

\usepackage[preprint]{acl}

\usepackage{times}
\usepackage{latexsym}

\usepackage[T1]{fontenc}

\usepackage[utf8]{inputenc}

\usepackage{microtype}

\usepackage{inconsolata}

\usepackage{graphicx}
\usepackage{booktabs}
\usepackage{multirow} 
\usepackage{amsmath}
\usepackage{float}
\usepackage{tabularx}
\usepackage{array}
\title{Cognitive Profiling of LRMs' Reasoning Traces Using Bloom’s Taxonomy}

\author{
  Maria-Eleni Zoumpoulidi$^{1}$ \quad
  Georgios Paraskevopoulos$^{1}$ \quad
  Alexandros Potamianos$^{2}$ \\
  $^{1}$Institute for Language and Speech Processing, Athena Research Center, Greece \\
  $^{2}$Speech and Language Processing Group, National Technical University of Athens, Greece \\
  \texttt{mzoumpoulidi@gmail.com} \quad
  \texttt{g.paraskevopoulos@athenarc.gr} \\
  \texttt{potam@central.ntua.gr}
}

\begin{document}
\maketitle
\begin{abstract}
Large Reasoning Models (LRMs) have revolutionized reasoning in LLMs, and the increasing public availability of reasoning traces creates valuable opportunities to study model behavior not only at the surface level but also at the granularity of individual reasoning steps. However, understanding the types of thinking employed during reasoning - which offers critical insights into models' reasoning patterns and enables actionable applications - remains underexplored. To address this gap, we introduce a framework for automatic annotation of reasoning steps through the lens of Bloom's Taxonomy, which classifies thinking into six cognitive levels, such as Remembering, Applying and Evaluating. Using this framework, we perform a large-scale analysis across models and datasets, revealing both similarities and differences in thinking patterns across models and tasks. Moreover, we demonstrate that thinking-type information derived from reasoning traces correlates with correctness, paving the way for improved reasoning. Our findings establish a fine-grained framework for analyzing thinking patterns in LRMs and provide actionable insights for enhancing reasoning quality.

\end{abstract}

\section{Introduction}
Reasoning has long been regarded as the cornerstone of human intellect, making its emulation one of the grand aspirations of language models.

Initial approaches relied on chain-of-thought (CoT) prompting, which encourages models to develop textual rationales step-by-step (\citet{NEURIPS2022_9d560961}, \citet{NEURIPS2022_8bb0d291}) and related variants (e.g., \citet{NEURIPS2023_271db992}, \citet{wang2023selfconsistency}). Recently, attention has shifted toward Large Reasoning Models (e.g. \citet{Guo_2025}), which use self-generated CoT rationales as training signals to perform explicit reasoning before generating final responses, thereby achieving enhanced performance. The increasing public availability of these reasoning traces prompted a growing body of research. Recognizing the value of step-level analysis for studying phenomena such as overthinking (\citet{kumar2026overthinkslowdownattacksreasoning}), where length does not guarantee improved correctness, recent research has evolved beyond initial surface-level statistical metrics such as accuracy to incorporate step-level analyses. For instance, \citet{marjanovic2026deepseekr} define a taxonomy of reasoning processes in DeepSeek-R1, revealing its core building blocks. Similarly, \citet{kargupta2026cognitive} propose a taxonomy of invariants, meta-cognitive controls, representations, and operations. \citet{li-etal-2025-understanding} and \citet{li2025schoenfeldsanatomymathematicalreasoning} analyze the reasoning steps of LLMs and LRMs through the lens of Schoenfeld's Episode Theory, a framework that decomposes reasoning for mathematical problem solving into procedural stages, called episodes, such as Reading, Planning, Implementation, Exploration, and Verification, revealing a consistent pattern in reasoning steps across reasoning models. 
Motivated by this line of work, we focus on a complementary yet underexplored question: rather than asking what process occurs at each reasoning step, we ask what type of thinking or cognitive function the model employs at each step.

\begin{figure*}[t]
\centering
\includegraphics[width=\textwidth]{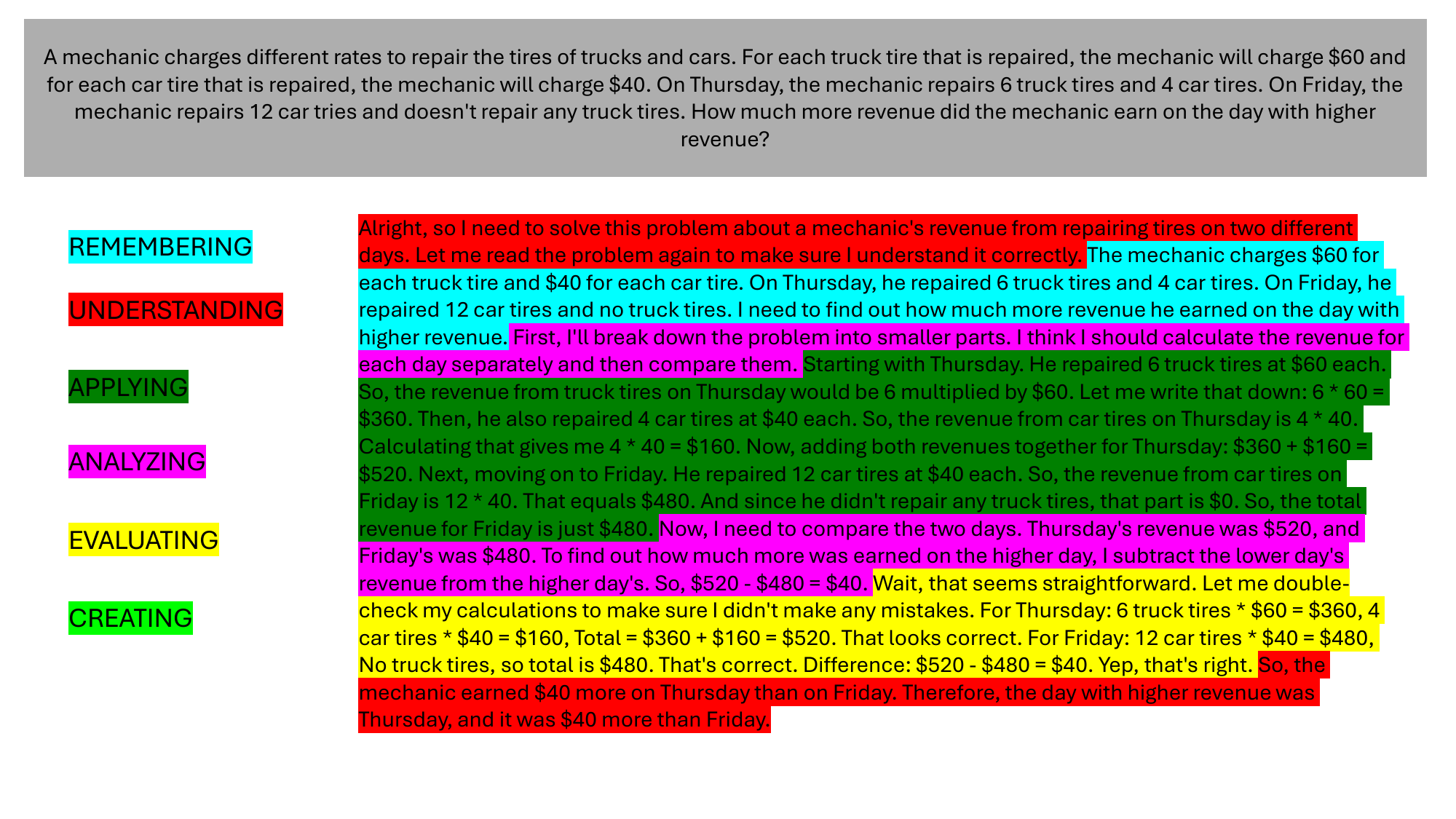}
\caption{Example of our annotation framework. The reasoning trace is automatically segmented, and each segment is labeled according to its corresponding level in Bloom’s Taxonomy.}
\label{fig:annotation_example}
\vspace{-1em}
\end{figure*}

We introduce a framework for analyzing LRM reasoning traces through the lens of Bloom’s Taxonomy, which hierarchically classifies cognition into six levels such as Remembering, Applying, and Evaluating (fig. ~\ref{fig:blooms_taxonomy}). We first generate chain-of-thought (CoT) traces, automatically segment them into reasoning steps, and annotate each step with a Bloom level using a human-validated automated pipeline. Figure ~\ref{fig:annotation_example} illustrates the annotation framework. Using these annotations, we analyze reasoning characteristics across diverse LRMs, datasets, and tasks to address the following research questions: 
\textbf{RQ1:} What similarities and differences exist in the cognitive profiles exhibited by different LRMs' internal reasoning traces for mathematical problem solving? \textbf{RQ2:} How do internal reasoning traces differ from output traces in cognitive profile? \textbf{RQ3:} How do cognitive profiles vary across different types of mathematical problems and among diverse (both mathematical and non-mathematical) tasks?
\textbf{RQ4:} Are Bloom-level features associated with correctness, and which cognitive transitions characterize correct/incorrect reasoning?

Our focus on cognitive function rather than process provides a complementary level of abstraction. For example, our framework distinguishes restating a question verbatim (Remembering) from constructing meaning through paraphrasing or inferring what is being asked (Understanding), whereas Schoenfeld classifies both as Reading. Conversely, we group recalling a relevant theorem or fact with recalling the question itself under Remembering, while Schoenfeld categorizes the former as Analyzing and the latter as Reading.

Our main contributions are as follows:
\begin{itemize}
    \item We introduce a novel framework for the automatic segmentation, annotation, and analysis of reasoning traces of LRMs through the lens of Bloom’s Taxonomy. Our code and data is available under the Apache 2.0 license \footnote{\url{{https://github.com/marilena1123/rlms_thinking_repo}}}.
    \item We analyze thinking patterns in LRM reasoning traces across models, datasets, and tasks, finding: for mathematical reasoning, (i) a shared remember–understand–apply–evaluate arc with model-specific traits, (ii) output trace cognitive compression relative to internal traces, and (iii) a shift from application to analysis with increasing difficulty; and across tasks, (iv) task-dependent profiles with back-loaded verification.
    \item We demonstrate the practical utility of our framework through a case study on correctness.
\end{itemize}
Our results yield novel insights into the thinking profiles of LRMs, revealing distinct patterns across models and tasks. Furthermore, they demonstrate the practical significance of thinking-level annotations for yielding improved reasoning.
\section{Related Work}
\subsection{Reasoning in LLMs}
Early research on LLMs emphasized eliciting reasoning through chain-of-thought (CoT) prompting (\citet{NEURIPS2022_9d560961}, \citet{NEURIPS2022_8bb0d291}), i.e., encouraging the model to produce step-by-step solutions, and related variants, such as traversing tree-like structures of reasoning states (Tree of Thoughts) \citep{NEURIPS2023_271db992} or sampling a diverse set of reasoning paths and then selecting the most consistent answer \citep{wang2023selfconsistency}. More recently, attention has shifted toward LRMs, where self-generated CoTs serve as training signals and reasoning mechanisms are more explicitly embedded within the model, enabling it to reason through a problem before producing a final answer. (e.g. \citet{Guo_2025}, \citet{qwen3technicalreport}). 

\subsection{Reasoning Traces Analysis}
A growing body of research aims to shed light on the reasoning traces of LRMs. Motivated by the value of step-level analysis for studying phenomena such as overthinking (\citet{kumar2026overthinkslowdownattacksreasoning}), where longer reasoning does not necessarily imply improved correctness, recent work has moved beyond surface-level statistical metrics such as accuracy toward more fine-grained step-level analyses. \citet{marjanovic2026deepseekr} define a taxonomy of reasoning processes in DeepSeek-R1, revealing its core building blocks: problem definition, followed by decomposition, and repeated reconstruction cycles before a final answer. They further examine whether reasoning chains correlate with human cognitive processes in terms of sentence processing load, and investigate factors such as reasoning length, context handling, and management of long or ambiguous inputs, as well as cultural and safety considerations. Similarly, \citet{kargupta2026cognitive} introduce a taxonomy covering reasoning invariants (e.g., logical coherence), meta-cognitive controls (e.g., self-awareness, strategy selection), representations (e.g., hierarchical), and operations (e.g., backtracking, forward chaining) to characterize the structural and procedural components of a reasoning trace. Grounding their analysis in Schoenfeld's Episode Theory- a framework that decomposes reasoning for mathematical problem solving into procedural stages, called episodes, such as Reading, Planning, Implementation, Exploration, and Verification, \citet{li-etal-2025-understanding} introduce an annotated corpus and automatic annotation framework for classifying reasoning behavior at the step level within reasoning traces, which \citet{li2025schoenfeldsanatomymathematicalreasoning} subsequently apply to large-scale comparative analysis across models, with both works focusing on mathematical problem solving. Their findings demonstrate a consistent heartbeat of reasoning steps across reasoning models. \citet{halim2025studythinkingpatternslarge} focus on reasoning traces in the context of code generation, identifying both human-like performance patterns and notable differences across models.

\subsection{Bloom's Taxonomy in NLP}
A growing body of NLP research employs Bloom's taxonomy. \citet{zoumpoulidi-etal-2025-bloomwise} introduce a Bloom’s taxonomy-inspired prompting technique aimed at enhancing mathematical reasoning. \citet{huber-niklaus-2025-llms} present a mapping of widely used benchmarks to Bloom’s taxonomy, identifying imbalances across cognitive levels. \citet{zoumpoulidi2025bloomxplain} introduce a framework for Bloom-aligned LLM-generated explanations. 

\section{Preliminaries: Bloom's Taxonomy}

\begin{figure}[H]
\centering
\includegraphics[width=0.7\linewidth]{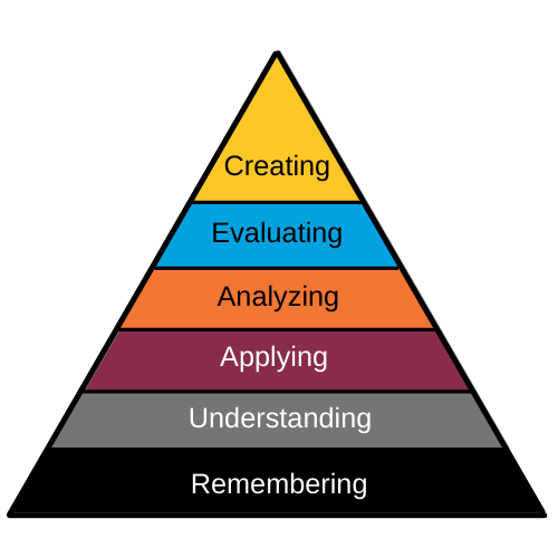}
\caption{Bloom's Taxonomy (revised by \citet{anderson2001taxonomy}).}
\label{fig:blooms_taxonomy}
\vspace{-1em}
\end{figure}

Bloom’s Taxonomy (fig. ~\ref{fig:blooms_taxonomy}) organizes thinking into a hierarchy of six levels of increasing complexity, defined as follows \citep{forehand2005blooms}:
\\
\noindent{\bf Remembering:} 
Retrieving, recognizing, and recalling relevant knowledge from long-term memory.\\
\noindent{\bf Understanding:} 
Constructing meaning from oral, written, and graphic messages through interpreting, exemplifying, classifying, summarizing, inferring, comparing, and explaining.\\
\noindent{\bf Applying:} 
Carrying out or using a procedure through executing, or implementing.\\
\noindent{\bf Analyzing:}
Breaking material into constituent parts, determining how the parts relate to one another and to an overall structure or purpose through differentiating, organizing, and attributing.\\
\noindent{\bf Evaluating:}
Making judgments based on criteria and standards through checking and critiquing.\\
\noindent{\bf Creating:}
Putting elements together to form a coherent or functional whole; reorganizing elements into a new pattern or structure through generating, planning, or producing.
\section{Our Approach}
\subsection{Methodology}
Our goal is to conduct a thorough analysis of the thinking patterns of LRMs' reasoning traces through the lens of Bloom’s Taxonomy. To this end, we design a multi-stage pipeline consisting of chain-of-thought (CoT) generation, step-level segmentation and annotation with Bloom’s Taxonomy labels, and subsequent analysis.

Our framework is structured as follows:
\begin{itemize}
    \item \textbf{CoT Generation:} We generate zero-shot chain-of-thought (CoT) solutions for each problem and store both the (internal) reasoning traces and the corresponding final responses. Full prompting details are provided in Appendix~\ref{sec:prompts}.
    \item \textbf{Annotation:} We employ \textit{Llama-3.3-70B-Instruct} to perform automatic annotation. Specifically, the model is prompted to segment each reasoning trace into discrete cognitive steps, assign a corresponding Bloom’s Taxonomy level to each step and provide a justification for the assigned label. We adopt LLM-based segmentation rather than more rigid alternatives, such as sentence-level annotation, because individual thinking processes may span multiple sentences. More structured approaches can therefore artificially fragment a single cognitive function, leading to less accurate annotations. This design choice reflects a trade-off between annotation uniformity and semantic fidelity. The full annotation prompt is provided in Appendix~\ref{sec:prompts}. To assess annotation reliability, two human annotators (one author and one non-author trained by the authors, both with an AI background and familiarity with Bloom's Taxonomy) independently labeled (based on the prompt) a uniformly randomly sampled subset of 100 samples drawn from all models and datasets, comprising 1,633 reasoning steps, and we measured agreement with the LLM-generated annotations (details in ~\ref{sec:human_annotation_details}). We compute Cohen’s Kappa  (Table~\ref{tab:agreement_comparison}), with the high agreement values indicating strong annotation quality. For simplicity, the evaluation focused solely on the labeling quality rather than on both segmentation and labeling. Segmentation quality was additionally assessed through qualitative inspection of 50 samples by one of the authors.
\begin{table}[H]
\centering
\small
\begin{tabular}{lccc}
\toprule
\textbf{Agreement} & \textbf{Cohen’s Kappa} \\
\midrule
H1-H2 &  0.8928   \\
LLM-H1 & 0.9173  \\
LLM-H2 & 0.9132   \\
\bottomrule
\end{tabular}
\caption{Cohen's Kappa between the LLM annotator (Llama3.3-70B-Instruct) and human annotators (H1, H2) on a uniformly randomly sampled subset of 100 samples (1633 reasoning steps) drawn from all models and datasets.}
\label{tab:agreement_comparison}
\end{table}
\end{itemize}
\subsection{Experimental setting}
\textbf{Datasets:} We conduct our main experiments on a diverse collection of three math reasoning datasets, each covering different challenging problem
types: GSM8K \citep{cobbe2021trainingverifierssolvemath}, GSM-hard \citep{DBLP:journals/corr/abs-2211-10435} and MATH500 (\citet{hendrycksmath2021}, \citet{ICLR2024_aca97732}). The GSM-hard dataset is a modified
version of GSM8K, where small numerical values have been replaced with larger ones to introduce greater computational difficulty. Together, these datasets span three levels of difficulty: grade-school (GSM8K), computationally intensive (GSM-hard), and competition-level (MATH500). The total number of samples is 3138 (GSM8K, GSM-hard:1319 each, MATH500: 500).
\\
\textbf{Models:} For our experiments, we query Qwen3-30B-A3B-Thinking-2507, Qwen3-4B-Thinking-2507 \citep{qwen3technicalreport}, DeepSeek-R1, R1-Distill-Qwen-1.5B, R1-Distill-Qwen-7B, DeepSeek-R1-Distill-Llama-8B \citep{Guo_2025} and Phi-4-reasoning \citep{abdin2025phi4reasoningtechnicalreport}. For all models, we use the default settings recommended by the providers to avoid introducing artifacts or distortions in the reasoning traces due to our experimental setup. For Phi-4-Reasoning, the default configuration includes a detailed system prompt, which is provided in the official report.
\begin{table*}[t]
\centering
\small
\begin{tabular}{lcccccc}
\toprule
\textbf{Model} & \textbf{Remember} & \textbf{Understand} & \textbf{Apply} & \textbf{Analyze} & \textbf{Evaluate} & \textbf{Create} \\
\midrule
DeepSeek-R1 & 11.8 & 24.0 & 32.9 & 12.5 & 18.6 & 0.2\\
DeepSeek-R1-Distill-Llama-8B & 13.5 & 18.7 & 44.2 & 14.1 & 9.3 & 0.3 \\
R1-Distill-Qwen-7B & 11.5 & 17.7 & 43.9 & 12.8 & 14.1 & 0.1  \\
R1-Distill-Qwen-1.5B & 10.9 & 18.8 & 35.6 & 13.5 & 21.0 & 0.1  \\
Qwen3-30B-A3B-Thinking-2507 & 9.4 & 18.8 & 33.4 & 12.7 & 25.6 & 0.1 \\
Qwen3-4B-Thinking-2507 & 9.0 & 20.8 & 32.5 & 14.3 & 23.1 & 0.2 \\
Phi-4-Reasoning & 28.4 & 21.0 & 26.9 & 10.1 & 12.7 & 0.9 \\
\bottomrule
\end{tabular}
\caption{Percentage of reasoning steps assigned to each level of Bloom’s Taxonomy across different LRMs.}
\label{tab:bloom_distribution_models}
\end{table*}
\section{Experiments}
\subsection{RQ1: Thinking Patterns Across LRMs in Mathematical Problem Solving}
\subsubsection{Bloom's Taxonomy Level Coverage}
Table~\ref{tab:bloom_distribution_models} reports the share of reasoning steps assigned to each Bloom's level across LRMs, considering only (internal) reasoning traces (output traces are excluded).
\paragraph{Apply dominates, Create is negligible.} Across all models, \textit{Applying} accounts for the largest or near-largest share of thinking steps (26.9\%--44.2\%), indicating that LRMs on mathematical tasks are predominantly oriented toward execution-type thinking. In contrast, \textit{Creating} is virtually absent in every model ($<$1.0\%), suggesting that LRMs rarely engage in the kind of thinking that synthesizes genuinely novel patterns or structures—consistent with the nature of the benchmarks used. \textit{Understanding} (17.7\%--24.0\%) and \textit{Analyzing} (10.1\%--14.3\%) occupy stable middle bands across all models, acting as cognitive connective tissue regardless of model family or scale.
\paragraph{Distillation profile is scale-dependent.} Comparing DeepSeek-R1 (32.9\% Apply, 18.6\% Evaluate) with its distilled variants reveals a systematic shift in thinking style: R1-Distill-Llama-8B (44.2\%, 9.3\%) and R1-Distill-Qwen-7B (43.9\%, 14.1\%) lean more heavily on execution-type thinking and engage less in evaluative thinking. This pattern is non-monotonic in scale: the smaller R1-Distill-Qwen-1.5B (35.6\%, 21.0\%) more closely resembles the teacher than its larger 7B and 8B siblings, suggesting that possibly the Apply-heavy bias intensifies as the distilled student grows.
\paragraph{The Qwen3 family is evaluation-heavy.} Both Qwen3-30B-A3B-Thinking-2507 (25.6\% Evaluate) and Qwen3-4B-Thinking-2507 (23.1\%) exhibit the highest proportions of \textit{Evaluating} thinking among all models, paired with the lowest proportions of \textit{Remembering} (9.0\%--9.4\%)—a cognitive style that favors checking and critiquing over fact recall.
\paragraph{Phi-4-Reasoning is a clear outlier.} Phi-4-Reasoning exhibits a markedly different cognitive profile: the highest share of \textit{Remembering} (28.4\%, more than double that of most other models), the lowest \textit{Analyzing} (10.1\%), the lowest \textit{Apply} (tied at 26.9\%), and the highest—though still small—share of \textit{Creating} (0.9\%). Rather than concentrating its thinking in \textit{Applying}, Phi-4-Reasoning distributes cognitive effort more evenly across levels, with an unusually large portion devoted to recall-type thinking. This may reflect differences in training data and post-training objectives relative to the DeepSeek and Qwen families.
\begin{figure*}[t]
\centering
\includegraphics[width=\textwidth]{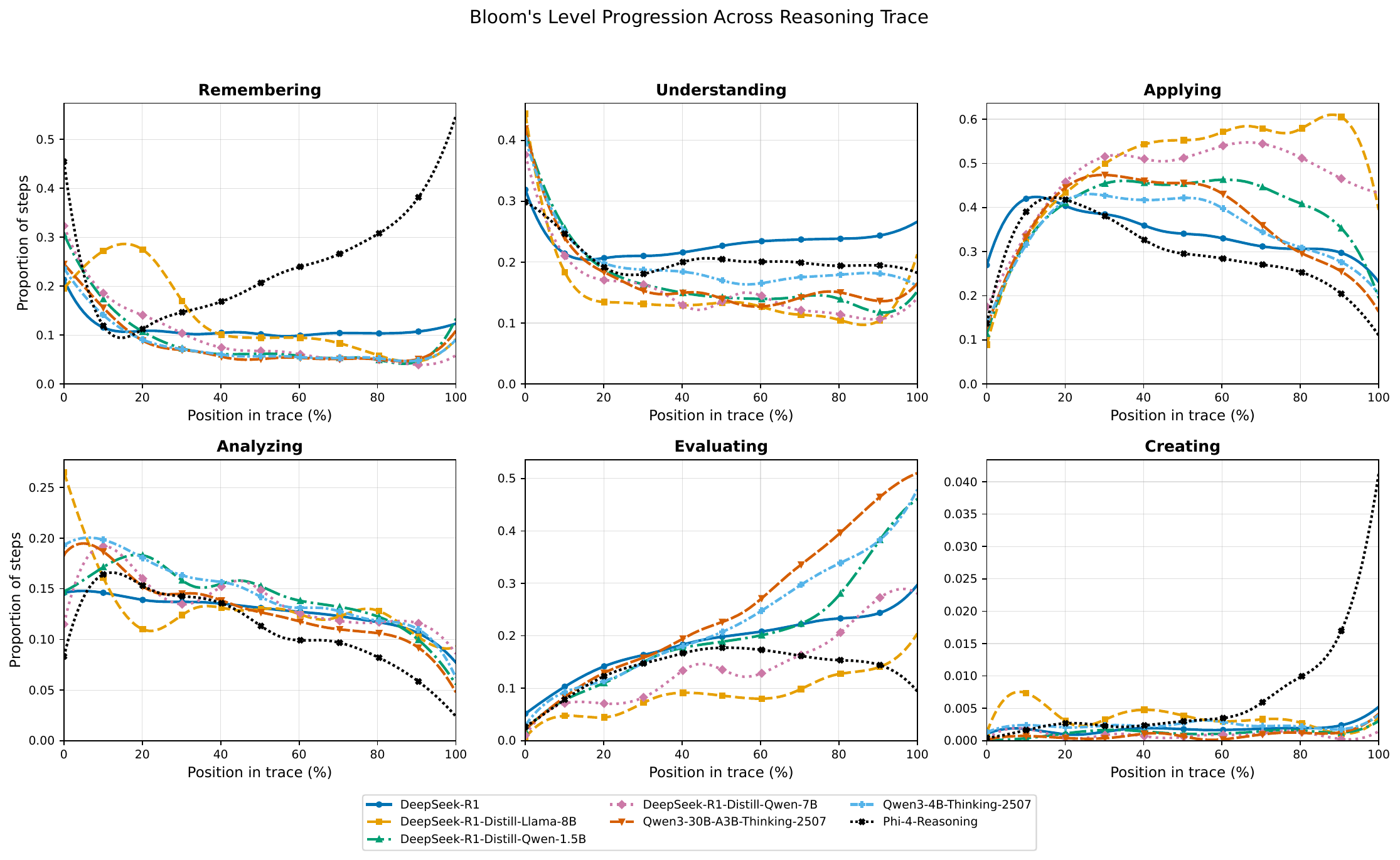}
\caption{Temporal Dynamics of Bloom's Levels in Reasoning Traces}
\label{fig:temporal_dynamics}
\vspace{-1em}
\end{figure*}
\subsubsection{Temporal Dynamics}
Figure~\ref{fig:temporal_dynamics} traces the proportion of each Bloom level as a function of relative position in the reasoning trace, revealing the temporal organization of thinking. For each trace, step positions are normalized to [0,100] and assigned to one of 10 equally spaced position bins. Each curve shows, within each bin, the share of steps at the corresponding Bloom level (again, output traces are excluded).

\paragraph{An emergent cognitive arc.} Across almost all models, the temporal dynamics of cognitive levels follow a coherent Remember-Understand-Apply-Evaluate arc that mirrors Bloom's hierarchy itself—an emergent ordering that, despite never being explicitly enforced, aligns with the canonical Bloom progression. We examine each level in turn.

\paragraph{Remembering is front-loaded—except for Phi-4-Reasoning.} For most models, \textit{Remembering} starts near 0.2--0.45 and decays sharply within the first 20\% of the trace, concentrating recall-type thinking at the outset. Notably, R1-Distill-Llama-8B, rises sharply before joining the general decay.  Some models show a slight late-trace rebound, suggesting a brief return to recall-type thinking near the end. Phi-4-Reasoning is the striking exception: its curve is U-shaped, rising sharply in the final portion to $\sim$0.55—significantly higher than that of any other model.

\paragraph{Understanding decays quickly but rebounds at the end.} \textit{Understanding} starts near 0.3--0.45 and collapses within the first 20\% of the trace for most models, mirroring \textit{Remembering}. Some models again show a slight late-trace rebound, suggesting a brief return to interpretation-type thinking near the final answer. DeepSeek-R1 stands out as the only model that sustains an elevated share throughout the entire trace, indicating a cognitive style that continues reinterpreting rather than fully committing to execution-type thinking.

\paragraph{Applying dominates the middle of the trace.} \textit{Applying} rises from roughly 0.1--0.3 to a sustained plateau of 0.4--0.6 across the middle of the trace, becoming the dominant mode of LRMs' thinking. The larger distilled R1 variants (Llama 8B and Qwen 7B) exhibit the highest peaks ($\sim$0.55--0.60), whereas DeepSeek-R1 and Phi-4-Reasoning maintain the lowest plateau ($\sim$0.42), which is reached relatively early, suggesting they transition more quickly into execution. Qwen models peak near the middle of the trace before declining, while the distilled models sustain elevated \textit{Applying} levels longer into the trace, indicating more persistent execution-type thinking.

\paragraph{Analyzing peaks early.} \textit{Analyzing} peaks slightly later than \textit{Remembering} and \textit{Understanding} (5--20\%) before gradually declining, consistent with early decomposition-type thinking. R1-Distill-Llama-8B shows an unusually high early share before dropping quickly, suggesting early decomposition into parts. Phi-4-Reasoning maintains the lowest proportion of \textit{Analyzing} steps in the middle and end of the trace.

\paragraph{Evaluating ramps up at the end---most aggressively for Qwen3 and the small distilled model.} \textit{Evaluating} mirrors \textit{Remembering}: near zero at the start and climbing monotonically by the end, reflecting back-loaded evaluative thinking. The two Qwen3 variants exhibit the steepest ramps ($\sim$0.48--0.52), with the smallest distilled variant (R1-Distill-Qwen-1.5B) following close behind ($\sim$0.46)---well above its larger 7B and 8B siblings and DeepSeek-R1. Phi-4-Reasoning again deviates, remaining flat and low throughout, with only a modest mid-trace peak.

\paragraph{Creating is negligible—except for a Phi-4-Reasoning spike at the end.} \textit{Creating} remains near zero throughout for all models. Phi-4-Reasoning is the sole exception, spiking at the end of the trace—small in absolute terms, but an order of magnitude above any other model. \\
For the interested reader, a further analysis of Phi-4-reasoning's divergent behavior (which we demonstrate to be a property of the model rather than its system prompt) can be found in Appendix ~\ref{sec:phi-analysis}.
\subsection{RQ2: Internal vs. Output Reasoning Trace Cognitive Profiles}
\begin{figure}[H]
\centering
\includegraphics[width=0.9\linewidth]{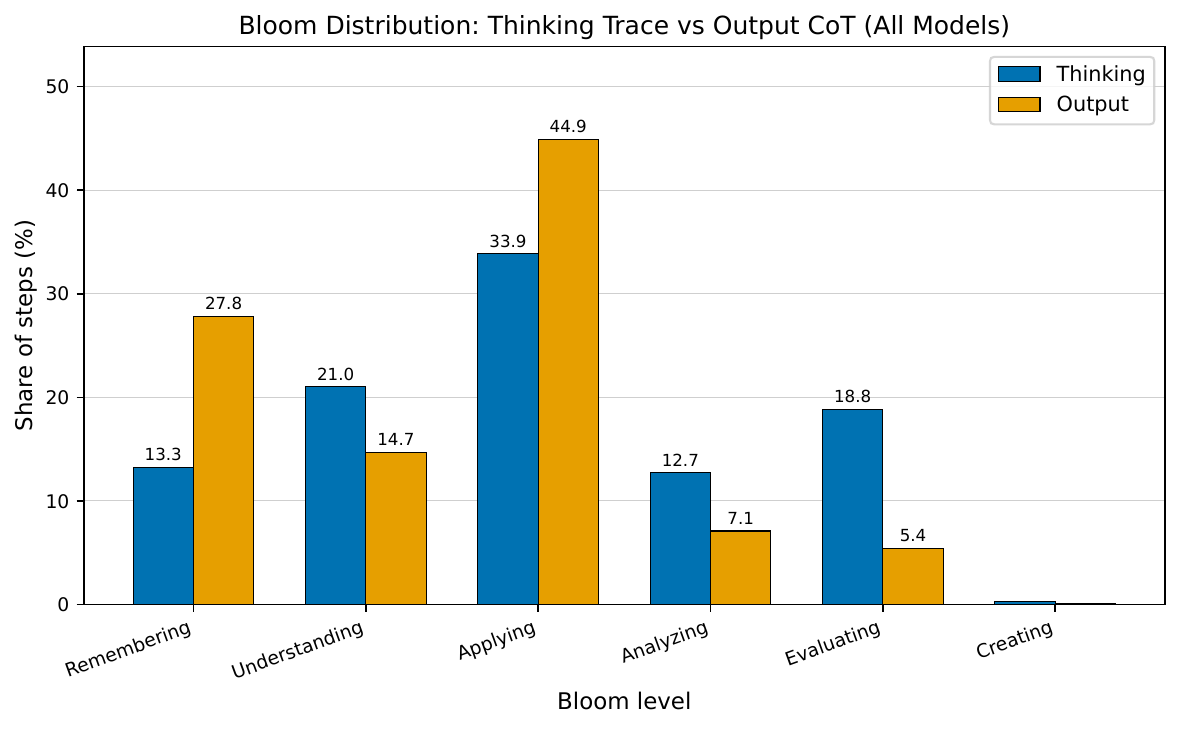}
\caption{Output-Reasoning steps distribution comparison}
\label{fig:output_vs_thinking_comparison}
\vspace{-1em}
\end{figure}

Figure~\ref{fig:output_vs_thinking_comparison} compares the thinking distribution of (internal) reasoning traces against final output CoTs, aggregated across all models. Thinking traces are cognitively broad, distributing effort across \textit{Applying} (33.9\%), \textit{Understanding} (21.0\%), \textit{Evaluating} (18.8\%), and \textit{Analyzing} (12.7\%). Outputs, by contrast, compress this richness into recall and execution: \textit{Remembering} (27.8\%) and \textit{Applying} (44.9\%) alone account for nearly three quarters of all steps, while evaluative and analytical thinking are largely stripped away (\textit{Evaluating} drops from 18.8\% to 5.4\%). This compression is expected and consistent with the role of final CoTs as concise step-by-step solutions, whereas internal reasoning traces more closely resemble think-aloud processes that include intermediate deliberation, revisiting of assumptions, evaluation, and exploratory reasoning.
\subsection{RQ3: Cognitive Profiles Across Tasks}
\subsubsection{Cognitive Profiles Across Mathematical Problem Types}
\begin{figure*}[t]
\centering
\includegraphics[width=\textwidth]{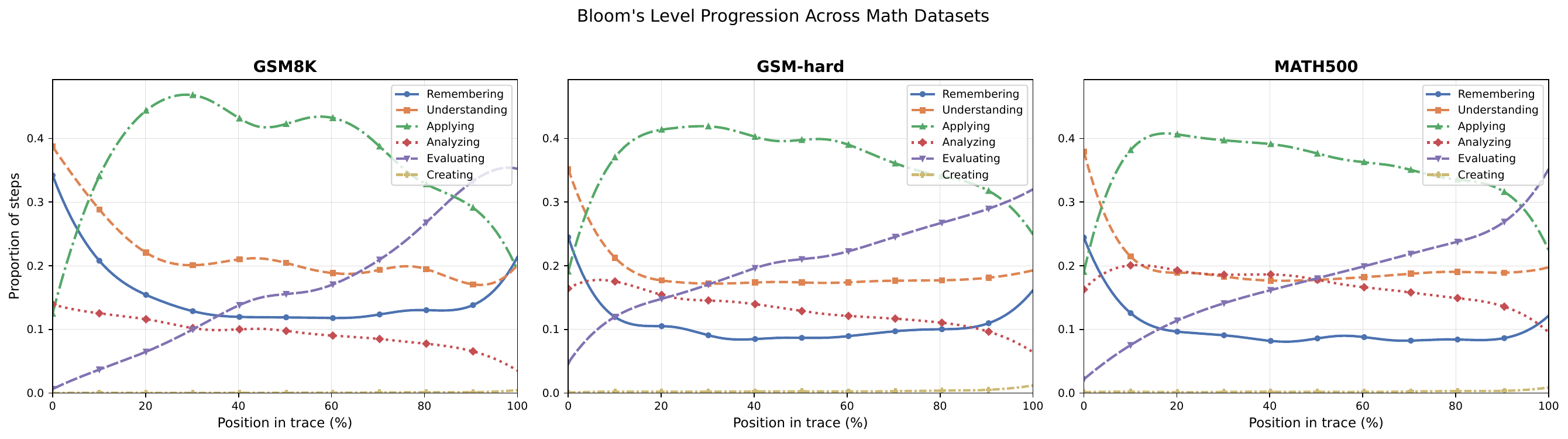}
\caption{Comparison of temporal thinking dynamics of reasoning traces across GSM8K, GSM-hard, and MATH500, aggregated over all models.}
\label{fig:math_datasets_bloom_traces}
\vspace{-1em}
\end{figure*}
Figure~\ref{fig:math_datasets_bloom_traces} shows the temporal thinking trajectories of reasoning traces across GSM8K, GSM-hard, and MATH500, aggregated over all models (output traces are excluded). Across GSM8K, GSM-hard, and MATH500, the overall remember--understand--apply--evaluate arc is preserved, but proportions shift with difficulty. \textit{Applying} dominates the middle of the trace in all three datasets, yet its plateau is highest on GSM8K ($\sim$0.47), flattens on GSM-hard ($\sim$0.40), and decays most steeply on MATH500, indicating that harder problems pull effort away from pure execution. \textit{Evaluating} ramps up toward the end across all datasets. \textit{Analyzing} is also more prominent on MATH500, sustaining an elevated share ($\sim$0.18--0.20) rather than decaying early as in GSM8K, consistent with the greater need for decomposition in competition-level problems. 
\subsubsection{Cognitive Profiles Across Diverse (mathematical and non-mathematical) Tasks}
\begin{figure*}[t]
\centering
\includegraphics[width=\textwidth]{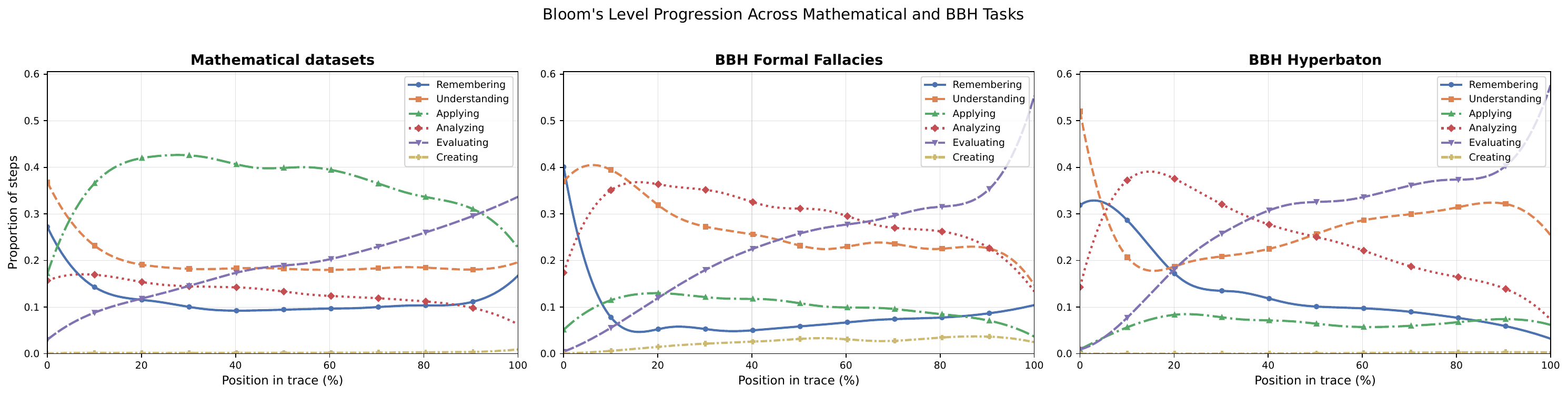}
\caption{Comparison of temporal thinking dynamics of reasoning traces across math datasets and the BBH tasks Formal Fallacies and Hyperbaton, aggregated over all models.}
\label{fig:task_variation}
\vspace{-1em}
\end{figure*}
Figure~\ref{fig:task_variation} compares the temporal thinking dynamics of reasoning traces  (aggregated over all models- internal traces only) across three reasoning domains: mathematical (pooled GSM8K, GSM-hard, and MATH500-all samples) and two BBH tasks \citep{suzgun-etal-2023-challenging}, 250 samples each — Formal Fallacies, which asks the model to assess the logical validity of informal arguments, and Hyperbaton, which requires determining the correct adjective ordering — capturing mathematical, logical, and linguistic reasoning respectively. The corresponding prompts and one-shot examples can be found in Appendix ~\ref{sec:prompts}. The profiles differ substantially. Mathematical reasoning is Apply-heavy, whereas Formal Fallacies emphasizes Analyzing and Understanding, reflecting the need to translate natural-language premises into logical structures and decompose their relations. Creating also rises slightly, consistent with generating counterexamples for validity testing. Hyperbaton shows a distinct pattern: Remembering and Understanding dominate early as the model interprets the question and recalls adjective-order rules, followed by an early Analyzing peak and sustained Understanding, while Applying remains consistently low, reflecting categorical rather than procedural reasoning. Both Formal Fallacies and Hyperbaton show elevated mid-trace Evaluating, reflecting the role of evaluation not only for verification but also in the task itself (validity or order assessment), while its late rise across tasks suggests back-loaded verification as a general LRM property. Examples are in Appendix~\ref{sec:example_outputs}.
\subsection{RQ4: Bloom-Level Features and Correctness}
To demonstrate the practical value of our framework, we investigate whether Bloom’s Taxonomy thinking levels are associated with solution correctness. We construct a balanced dataset for the mathematical datasets by independently balancing each model $\times$ dataset cell, by retaining all samples from the minority correctness class and randomly sampling an equal number from the majority class (9132 samples in total). 
We then train an $L_1$-regularized logistic regression model to predict correctness from 43 standardized features: total token count, six Bloom-level token proportions, and the flattened $6 \times 6$ transition matrix normalized by the number of transitions. Evaluation uses nested cross-validation with a 5-fold outer loop and a 3-fold inner loop. Coefficients can be found in Table ~\ref{tab:bloom_coeffs}. Applying $\rightarrow$ evaluating is the strongest positive coefficient, suggesting that correct solutions often move from execution-oriented reasoning into explicit assessment of intermediate results. Remembering $\rightarrow$ understanding also receives a strong positive coefficient, indicating that successful traces frequently benefit from explicitly constructing meaning from recalled information. By contrast, total token count is the strongest predictor of incorrectness, indicating that longer traces are generally associated with lower solution quality. Transitions such as evaluating $\rightarrow$ analyzing and analyzing $\rightarrow$ evaluating suggest that unsuccessful traces more frequently oscillate between analysis and evaluation without converging on a stable solution. Overall, incorporating Bloom-level proportions and transition dynamics improves performance over a length-only baseline, where the only feature is the number of tokens. The full 43-feature model achieves an AUC of $0.676 \pm 0.012$ compared to $0.613 \pm 0.008$ for the baseline, with smaller gains in accuracy ($0.631$ vs.\ $0.623$) and F1 ($0.667$ vs.\ $0.666$). These findings suggest that Bloom-based features capture information related to correctness which have the potential to yield improved reasoning by explicitly shaping such structural patterns during training or inference.

\begin{table}[t]
\centering
\resizebox{\columnwidth}{!}{
\begin{tabular}{lr|lr}
\toprule
\multicolumn{2}{c|}{\textbf{Positive Coefficients}} & 
\multicolumn{2}{c}{\textbf{Negative Coefficients}} \\
\textbf{Feature} & \textbf{Coef.} & \textbf{Feature} & \textbf{Coef.} \\
\midrule
trans\_appl$\rightarrow$eval & +0.1769 & total\_tokens & -0.4373 \\
trans\_reme$\rightarrow$unde & +0.1290 & trans\_eval$\rightarrow$anal & -0.1271 \\
trans\_eval$\rightarrow$reme & +0.1059 & trans\_anal$\rightarrow$eval & -0.1236 \\
trans\_eval$\rightarrow$eval & +0.0897 & trans\_eval$\rightarrow$appl & -0.1155 \\
trans\_appl$\rightarrow$unde & +0.0813 & trans\_eval$\rightarrow$crea & -0.1132 \\
trans\_eval$\rightarrow$unde & +0.0598 & tok\_analyzing & -0.0677 \\
trans\_reme$\rightarrow$eval & +0.0568 & trans\_unde$\rightarrow$appl & -0.0559 \\
trans\_appl$\rightarrow$reme & +0.0473 & & \\
\bottomrule
\end{tabular}
}
\caption{Top positive and negative logistic regression coefficients from the Bloom-based correctness model. Positive coefficients are associated with correct solutions, while negative coefficients are associated with incorrect solutions.}
\label{tab:bloom_coeffs}
\end{table}
\section{Conclusions}
We introduce a framework for automatically annotating LRM reasoning steps using Bloom’s Taxonomy. Analysis across LRMs and datasets reveals: for mathematical reasoning, (i) a shared remember–understand–apply–evaluate arc with model-specific traits; (ii) internal traces are cognitively broader than output CoTs, which compress into recall and execution; (iii) cognitive effort shifts from execution to analysis with increasing difficulty; across tasks, (iv) cognitive profiles are task-dependent, though back-loaded verification remains consistent; and (v) Bloom-based features correlate with correctness, suggesting potential for improving reasoning.

\section*{Limitations}
We acknowledge that, despite the informative and actionable insights provided by our method and analysis - particularly regarding the relationship between correctness and reasoning depth - this work does not explicitly enforce or shape such structural patterns to improve reasoning performance, which we plan to explore in future work. Additionally, extending the study to a broader and more diverse set of tasks would enable a more comprehensive and well-rounded analysis.

\bibliography{custom}

@inproceedings{NEURIPS2022_9d560961,
 author = {Wei, Jason and Wang, Xuezhi and Schuurmans, Dale and Bosma, Maarten and ichter, brian and Xia, Fei and Chi, Ed and Le, Quoc V and Zhou, Denny},
 booktitle = {Advances in Neural Information Processing Systems},
 editor = {S. Koyejo and S. Mohamed and A. Agarwal and D. Belgrave and K. Cho and A. Oh},
 pages = {24824--24837},
 publisher = {Curran Associates, Inc.},
 title = {Chain-of-Thought Prompting Elicits Reasoning in Large Language Models},
 url = {https://proceedings.neurips.cc/paper_files/paper/2022/file/9d5609613524ecf4f15af0f7b31abca4-Paper-Conference.pdf},
 volume = {35},
 year = {2022}
}

@inproceedings{NEURIPS2022_8bb0d291,
 author = {Kojima, Takeshi and Gu, Shixiang (Shane) and Reid, Machel and Matsuo, Yutaka and Iwasawa, Yusuke},
 booktitle = {Advances in Neural Information Processing Systems},
 editor = {S. Koyejo and S. Mohamed and A. Agarwal and D. Belgrave and K. Cho and A. Oh},
 pages = {22199--22213},
 publisher = {Curran Associates, Inc.},
 title = {Large Language Models are Zero-Shot Reasoners},
 url = {https://proceedings.neurips.cc/paper_files/paper/2022/file/8bb0d291acd4acf06ef112099c16f326-Paper-Conference.pdf},
 volume = {35},
 year = {2022}
}

@inproceedings{NEURIPS2023_271db992,
 author = {Yao, Shunyu and Yu, Dian and Zhao, Jeffrey and Shafran, Izhak and Griffiths, Tom and Cao, Yuan and Narasimhan, Karthik},
 booktitle = {Advances in Neural Information Processing Systems},
 editor = {A. Oh and T. Naumann and A. Globerson and K. Saenko and M. Hardt and S. Levine},
 pages = {11809--11822},
 publisher = {Curran Associates, Inc.},
 title = {Tree of Thoughts: Deliberate Problem Solving with Large Language Models},
 url = {https://proceedings.neurips.cc/paper_files/paper/2023/file/271db9922b8d1f4dd7aaef84ed5ac703-Paper-Conference.pdf},
 volume = {36},
 year = {2023}
}

@inproceedings{zoumpoulidi-etal-2025-bloomwise,
    title = "{B}loom{W}ise: Enhancing Problem-Solving capabilities of Large Language Models using Bloom{'}s-Taxonomy-Inspired Prompts",
    author = "Zoumpoulidi, Maria-Eleni  and
      Paraskevopoulos, Georgios  and
      Potamianos, Alexandros",
    editor = "Valentino, Marco  and
      Ferreira, Deborah  and
      Thayaparan, Mokanarangan  and
      Ranaldi, Leonardo  and
      Freitas, Andre",
    booktitle = "Proceedings of The 3rd Workshop on Mathematical Natural Language Processing (MathNLP 2025)",
    month = nov,
    year = "2025",
    address = "Suzhou, China",
    publisher = "Association for Computational Linguistics",
    url = "https://aclanthology.org/2025.mathnlp-main.3/",
    doi = "10.18653/v1/2025.mathnlp-main.3",
    pages = "34--49",
    ISBN = "979-8-89176-348-7"
}

@inproceedings{
zoumpoulidi2025bloomxplain,
title={BloomXplain: A Framework and Benchmark Dataset for Pedagogically Sound {LLM}-Generated Explanations Based on Bloom{\textquoteright}s Taxonomy},
author={Maria-Eleni Zoumpoulidi and Eleni Batsi and Georgios Paraskevopoulos and Vassilis Katsouros and Alexandros Potamianos},
booktitle={NeurIPS 2025 Workshop on Evaluating the Evolving LLM Lifecycle: Benchmarks, Emergent Abilities, and Scaling},
year={2025},
url={https://openreview.net/forum?id=EMMPng2LLl}
}

@book{anderson2001taxonomy,
  author    = {L. W. Anderson and D. R. Krathwohl},
  title     = {A Taxonomy for Learning, Teaching, and Assessing: A Revision of Bloom’s Taxonomy of Educational Objectives},
  year      = {2001},
  publisher = {Longman},
}

@article{forehand2005blooms,
  title={Bloom's taxonomy: Original and revised},
  author={Forehand, Mary and others},
  journal={Emerging Perspectives on Learning, Teaching, and Technology},
  year={2005},
  volume={8},
  pages={41--44}
}

@inproceedings{
wang2023selfconsistency,
title={Self-Consistency Improves Chain of Thought Reasoning in Language Models},
author={Xuezhi Wang and Jason Wei and Dale Schuurmans and Quoc V Le and Ed H. Chi and Sharan Narang and Aakanksha Chowdhery and Denny Zhou},
booktitle={The Eleventh International Conference on Learning Representations },
year={2023},
url={https://openreview.net/forum?id=1PL1NIMMrw}
}

@misc{cobbe2021trainingverifierssolvemath,
      title={Training Verifiers to Solve Math Word Problems}, 
      author={Karl Cobbe and Vineet Kosaraju and Mohammad Bavarian and Mark Chen and Heewoo Jun and Lukasz Kaiser and Matthias Plappert and Jerry Tworek and Jacob Hilton and Reiichiro Nakano and Christopher Hesse and John Schulman},
      year={2021},
      eprint={2110.14168},
      archivePrefix={arXiv},
      primaryClass={cs.LG},
      url={https://arxiv.org/abs/2110.14168}, 
}

@article{DBLP:journals/corr/abs-2211-10435,
    author = {Luyu Gao and Aman Madaan and Shuyan Zhou and Uri Alon and Pengfei Liu and Yiming Yang and Jamie Callan and Graham Neubig},
    title = {{PAL: Program-Aided Language Models}},
    journal = {CoRR},
    volume = {abs/2211.10435},
    year = {2022},
    url = {https://doi.org/10.48550/arXiv.2211.10435},
    note = {Preprint available at arXiv}
}

@article{hendrycksmath2021,
  title={Measuring Mathematical Problem Solving With the MATH Dataset},
  author={Dan Hendrycks and Collin Burns and Saurav Kadavath and Akul Arora and Steven Basart and Eric Tang and Dawn Song and Jacob Steinhardt},
  journal={NeurIPS},
  year={2021}
}

@misc{qwen3technicalreport,
      title={Qwen3 Technical Report}, 
      author={Qwen Team},
      year={2025},
      eprint={2505.09388},
      archivePrefix={arXiv},
      primaryClass={cs.CL},
      url={https://arxiv.org/abs/2505.09388}, 
}

@article{
marjanovic2026deepseekr,
title={DeepSeek-R1 Thoughtology: Let{\textquoteright}s think about {LLM} reasoning},
author={Sara Vera Marjanovic and Arkil Patel and Vaibhav Adlakha and Milad Aghajohari and Parishad BehnamGhader and Mehar Bhatia and Aditi Khandelwal and Austin Kraft and Benno Krojer and Xing Han L{\`u} and Nicholas Meade and Dongchan Shin and Amirhossein Kazemnejad and Gaurav Kamath and Marius Mosbach and Karolina Stanczak and Siva Reddy},
journal={Transactions on Machine Learning Research},
issn={2835-8856},
year={2026},
url={https://openreview.net/forum?id=BZwKsiRnJI},
note={}
}

@article{Guo_2025,
   title={DeepSeek-R1 incentivizes reasoning in LLMs through reinforcement learning},
   volume={645},
   ISSN={1476-4687},
   url={http://dx.doi.org/10.1038/s41586-025-09422-z},
   DOI={10.1038/s41586-025-09422-z},
   number={8081},
   journal={Nature},
   publisher={Springer Science and Business Media LLC},
   author={Guo, Daya and Yang, Dejian and Zhang, Haowei and Song, Junxiao and Wang, Peiyi and Zhu, Qihao and Xu, Runxin and Zhang, Ruoyu and Ma, Shirong and Bi, Xiao and Zhang, Xiaokang and Yu, Xingkai and Wu, Yu and Wu, Z. F. and Gou, Zhibin and Shao, Zhihong and Li, Zhuoshu and Gao, Ziyi and Liu, Aixin and Xue, Bing and Wang, Bingxuan and Wu, Bochao and Feng, Bei and Lu, Chengda and Zhao, Chenggang and Deng, Chengqi and Ruan, Chong and Dai, Damai and Chen, Deli and Ji, Dongjie and Li, Erhang and Lin, Fangyun and Dai, Fucong and Luo, Fuli and Hao, Guangbo and Chen, Guanting and Li, Guowei and Zhang, H. and Xu, Hanwei and Ding, Honghui and Gao, Huazuo and Qu, Hui and Li, Hui and Guo, Jianzhong and Li, Jiashi and Chen, Jingchang and Yuan, Jingyang and Tu, Jinhao and Qiu, Junjie and Li, Junlong and Cai, J. L. and Ni, Jiaqi and Liang, Jian and Chen, Jin and Dong, Kai and Hu, Kai and You, Kaichao and Gao, Kaige and Guan, Kang and Huang, Kexin and Yu, Kuai and Wang, Lean and Zhang, Lecong and Zhao, Liang and Wang, Litong and Zhang, Liyue and Xu, Lei and Xia, Leyi and Zhang, Mingchuan and Zhang, Minghua and Tang, Minghui and Zhou, Mingxu and Li, Meng and Wang, Miaojun and Li, Mingming and Tian, Ning and Huang, Panpan and Zhang, Peng and Wang, Qiancheng and Chen, Qinyu and Du, Qiushi and Ge, Ruiqi and Zhang, Ruisong and Pan, Ruizhe and Wang, Runji and Chen, R. J. and Jin, R. L. and Chen, Ruyi and Lu, Shanghao and Zhou, Shangyan and Chen, Shanhuang and Ye, Shengfeng and Wang, Shiyu and Yu, Shuiping and Zhou, Shunfeng and Pan, Shuting and Li, S. S. and Zhou, Shuang and Wu, Shaoqing and Yun, Tao and Pei, Tian and Sun, Tianyu and Wang, T. and Zeng, Wangding and Liu, Wen and Liang, Wenfeng and Gao, Wenjun and Yu, Wenqin and Zhang, Wentao and Xiao, W. L. and An, Wei and Liu, Xiaodong and Wang, Xiaohan and Chen, Xiaokang and Nie, Xiaotao and Cheng, Xin and Liu, Xin and Xie, Xin and Liu, Xingchao and Yang, Xinyu and Li, Xinyuan and Su, Xuecheng and Lin, Xuheng and Li, X. Q. and Jin, Xiangyue and Shen, Xiaojin and Chen, Xiaosha and Sun, Xiaowen and Wang, Xiaoxiang and Song, Xinnan and Zhou, Xinyi and Wang, Xianzu and Shan, Xinxia and Li, Y. K. and Wang, Y. Q. and Wei, Y. X. and Zhang, Yang and Xu, Yanhong and Li, Yao and Zhao, Yao and Sun, Yaofeng and Wang, Yaohui and Yu, Yi and Zhang, Yichao and Shi, Yifan and Xiong, Yiliang and He, Ying and Piao, Yishi and Wang, Yisong and Tan, Yixuan and Ma, Yiyang and Liu, Yiyuan and Guo, Yongqiang and Ou, Yuan and Wang, Yuduan and Gong, Yue and Zou, Yuheng and He, Yujia and Xiong, Yunfan and Luo, Yuxiang and You, Yuxiang and Liu, Yuxuan and Zhou, Yuyang and Zhu, Y. X. and Huang, Yanping and Li, Yaohui and Zheng, Yi and Zhu, Yuchen and Ma, Yunxian and Tang, Ying and Zha, Yukun and Yan, Yuting and Ren, Z. Z. and Ren, Zehui and Sha, Zhangli and Fu, Zhe and Xu, Zhean and Xie, Zhenda and Zhang, Zhengyan and Hao, Zhewen and Ma, Zhicheng and Yan, Zhigang and Wu, Zhiyu and Gu, Zihui and Zhu, Zijia and Liu, Zijun and Li, Zilin and Xie, Ziwei and Song, Ziyang and Pan, Zizheng and Huang, Zhen and Xu, Zhipeng and Zhang, Zhongyu and Zhang, Zhen},
   year={2025},
   month={September}, pages={633–638} }

@misc{halim2025studythinkingpatternslarge,
      title={A Study on Thinking Patterns of Large Reasoning Models in Code Generation}, 
      author={Kevin Halim and Sin G. Teo and Ruitao Feng and Zhenpeng Chen and Yang Gu and Chong Wang and Yang Liu},
      year={2025},
      eprint={2509.13758},
      archivePrefix={arXiv},
      primaryClass={cs.SE},
      url={https://arxiv.org/abs/2509.13758}, 
}

@inproceedings{huber-niklaus-2025-llms,
    title = "{LLM}s meet Bloom{'}s Taxonomy: A Cognitive View on Large Language Model Evaluations",
    author = "Huber, Thomas  and
      Niklaus, Christina",
    editor = "Rambow, Owen  and
      Wanner, Leo  and
      Apidianaki, Marianna  and
      Al-Khalifa, Hend  and
      Eugenio, Barbara Di  and
      Schockaert, Steven",
    booktitle = "Proceedings of the 31st International Conference on Computational Linguistics",
    month = jan,
    year = "2025",
    address = "Abu Dhabi, UAE",
    publisher = "Association for Computational Linguistics",
    url = "https://aclanthology.org/2025.coling-main.350/",
    pages = "5211--5246"
}

@inproceedings{li-etal-2025-understanding,
    title = "Understanding the Thinking Process of Reasoning Models: A Perspective from Schoenfeld{'}s Episode Theory",
    author = "Li, Ming  and
      Zhang, Nan  and
      Fan, Chenrui  and
      Jiao, Hong  and
      Fu, Yanbin  and
      Peters, Sydney  and
      Xu, Qingshu  and
      Lissitz, Robert  and
      Zhou, Tianyi",
    editor = "Christodoulopoulos, Christos  and
      Chakraborty, Tanmoy  and
      Rose, Carolyn  and
      Peng, Violet",
    booktitle = "Proceedings of the 2025 Conference on Empirical Methods in Natural Language Processing",
    month = nov,
    year = "2025",
    address = "Suzhou, China",
    publisher = "Association for Computational Linguistics",
    url = "https://aclanthology.org/2025.emnlp-main.922/",
    doi = "10.18653/v1/2025.emnlp-main.922",
    pages = "18267--18288",
    ISBN = "979-8-89176-332-6"
}

@misc{li2025schoenfeldsanatomymathematicalreasoning,
      title={Schoenfeld's Anatomy of Mathematical Reasoning by Language Models}, 
      author={Ming Li and Chenrui Fan and Yize Cheng and Soheil Feizi and Tianyi Zhou},
      year={2025},
      eprint={2512.19995},
      archivePrefix={arXiv},
      primaryClass={cs.CL},
      url={https://arxiv.org/abs/2512.19995}, 
}

@misc{abdin2025phi4reasoningtechnicalreport,
      title={Phi-4-reasoning Technical Report}, 
      author={Marah Abdin and Sahaj Agarwal and Ahmed Awadallah and Vidhisha Balachandran and Harkirat Behl and Lingjiao Chen and Gustavo de Rosa and Suriya Gunasekar and Mojan Javaheripi and Neel Joshi and Piero Kauffmann and Yash Lara and Caio César Teodoro Mendes and Arindam Mitra and Besmira Nushi and Dimitris Papailiopoulos and Olli Saarikivi and Shital Shah and Vaishnavi Shrivastava and Vibhav Vineet and Yue Wu and Safoora Yousefi and Guoqing Zheng},
      year={2025},
      eprint={2504.21318},
      archivePrefix={arXiv},
      primaryClass={cs.AI},
      url={https://arxiv.org/abs/2504.21318}, 
}

@misc{kumar2026overthinkslowdownattacksreasoning,
      title={OverThink: Slowdown Attacks on Reasoning LLMs}, 
      author={Abhinav Kumar and Jaechul Roh and Ali Naseh and Marzena Karpinska and Mohit Iyyer and Amir Houmansadr and Eugene Bagdasarian},
      year={2026},
      eprint={2502.02542},
      archivePrefix={arXiv},
      primaryClass={cs.LG},
      url={https://arxiv.org/abs/2502.02542}, 
}

@inproceedings{suzgun-etal-2023-challenging,
    title = "Challenging {BIG}-Bench Tasks and Whether Chain-of-Thought Can Solve Them",
    author = {Suzgun, Mirac  and
      Scales, Nathan  and
      Sch{\"a}rli, Nathanael  and
      Gehrmann, Sebastian  and
      Tay, Yi  and
      Chung, Hyung Won  and
      Chowdhery, Aakanksha  and
      Le, Quoc  and
      Chi, Ed  and
      Zhou, Denny  and
      Wei, Jason},
    editor = "Rogers, Anna  and
      Boyd-Graber, Jordan  and
      Okazaki, Naoaki",
    booktitle = "Findings of the Association for Computational Linguistics: ACL 2023",
    month = jul,
    year = "2023",
    address = "Toronto, Canada",
    publisher = "Association for Computational Linguistics",
    url = "https://aclanthology.org/2023.findings-acl.824/",
    doi = "10.18653/v1/2023.findings-acl.824",
    pages = "13003--13051"
}

@inproceedings{
kargupta2026cognitive,
title={Cognitive Foundations for Reasoning and Their Manifestation in {LLM}s},
author={Priyanka Kargupta and Shuyue Stella Li and Haocheng Wang and Dean Light and Jinu Lee and Shan Chen and Orevaoghene Ahia and Kerem Oktar and Thomas L. Griffiths and Max Kleiman-Weiner and Jiawei Han and Asli Celikyilmaz and Yulia Tsvetkov},
booktitle={ICLR 2026 Workshop - From Human Cognition to AI Reasoning: Models, Methods, and Applications},
year={2026},
url={https://openreview.net/forum?id=rRSiiKAxCx}
}

@inproceedings{ICLR2024_aca97732,
 author = {Lightman, Hunter and Kosaraju, Vineet and Burda, Yuri and Edwards, Harrison and Baker, Bowen and Lee, Teddy and Leike, Jan and Schulman, John  and Sutskever, Ilya and Cobbe, Karl},
 booktitle = {International Conference on Learning Representations},
 editor = {B. Kim and Y. Yue and S. Chaudhuri and K. Fragkiadaki and M. Khan and Y. Sun},
 pages = {39578--39601},
 title = {Let\textquotesingle s Verify Step by Step},
 url = {https://proceedings.iclr.cc/paper_files/paper/2024/file/aca97732e30bcf1303bc22ac3924fd16-Paper-Conference.pdf},
 volume = {2024},
 year = {2024}
}

\appendix

\section{Prompts}
\label{sec:prompts}
In this section, we provide the prompts used for CoT generation (Mathematical Datasets: Table ~\ref{tab:cot_prompt}, BBH tasks: ~\ref{tab:cot_prompt_bbh}) and automatic segmentation of reasoning traces into discrete cognitive steps and assignment of a corresponding Bloom’s Taxonomy level to each step with a justification (Table ~\ref{tab:annotation_prompt}). The few-shot examples used for the annotation of the mathematical datasets can be found in Table ~\ref{tab:annotation_few_shpt_examples} and examples used for BBH formal fallacies and hyperbaton can be found in Tables ~\ref{tab:annotation_few_shpt_example_bbh} and ~\ref{tab:annotation_few_shpt_example_bbh_hyperbaton} respectively.
\begin{table*}[t]
\centering
\small
\begin{tabular}{p{0.95\linewidth}}
\hline
\textbf{CoT Prompt Template (Mathematical Datasets)} \\ 
\hline
Solve the following math problem step by step. End your response with the final numerical answer in this format: ``The answer is X.'' \\[4pt]

Problem: \texttt{\{question\}} \\
\hline
\end{tabular}
\caption{Prompt template used for Chain of Thought generation (Mathematical Datasets)}
\label{tab:cot_prompt}
\end{table*}

\begin{table*}[t]
\centering
\small
\begin{tabular}{p{0.95\linewidth}}
\hline
\textbf{Annotation Prompt Template} \\ 
\hline
You are a data annotation expert.

Given: \\
- A question \\
- A reasoning trace

Your task is to segment the reasoning trace into distinct steps, each aligned with exactly one cognitive level as defined by Bloom's Taxonomy.

\#\# Segmentation and Labeling Rules \\
1. Segment by cognitive function, not by sentence. A segment can span multiple sentences or be a single clause — boundaries are determined by shifts in cognitive operation, not punctuation. \\
2. Start a new segment whenever the solver completes a cognitive operation (e.g., stops recalling and starts applying, or stops applying and starts evaluating). This can include consecutive operations of the same type. \\
3. Transitional phrases ("Now...", "So...", "Let me...") belong to the segment of the operation they introduce, not the one before. \\
4. Higher taxonomy levels can subsume lower ones (e.g., Evaluate may involve Apply internally). In that case, label the segment by the highest level that characterizes the overall operation. \\
5. KEEP THE TEXT VERBATIM. DO NOT IMPROVE OR ALTER IT IN ANY WAY. DO NOT ADD text that is not present. Even when the reasoning is brief or incorrect, annotate it WITHOUT ADDITIONS and BASED ON WHAT IS PRESENT WITHOUT MAKING ANY ASSUMPTIONS ABOUT WHAT IS IMPLIED. OUR GOAL IS TO ANNOTATE THE EXISTING REASONING EXACTLY AS-IS. Do not omit any part of the reasoning trace. \\

\#\# Bloom's Taxonomy Labels \\
Assign each segment exactly one label:\\

- REMEMBER — Retrieves, recognizes, recalls or restates a fact, number, rule, or any information directly from the problem or prior knowledge, with no transformation or interpretation.\\
- UNDERSTAND — Constructs meaning from information: interprets, infers, summarizes, compares, classifies, exemplifies, or explains what the information means or implies — going beyond repetition but not yet acting on it.\\
- APPLY — Executes or implements a procedure, method, or algorithm. \\
- ANALYZE — Breaks the problem or information into parts, determines how those parts relate to each other and to the overall goal, or identifies dependencies between quantities and operations. \\
- EVALUATE — Makes a judgment based on criteria: checks whether a result is correct, verifies whether a method was valid, or critiques the quality of an approach. \\
- CREATE — Combines or synthesizes elements to produce a novel structure, formula, or solution strategy not derivable from a standard procedure. Assign only if a genuinely new path is invented. \\
Few-shot examples: [omitted due to space constraints, please see the corresponding tables] \\
Question \\ 
\texttt{\{question\}} \\[0.5em]

Reasoning Text to Segment and Classify \\ 
\texttt{\{reasoning\_text\}} \\[0.5em]

Output Format \\

One line per segment, in the order the reasoning proceeds. \\
Copy each segment's text verbatim --- do not paraphrase or omit any part. \\
The concatenation of all step texts must exactly reconstruct the full reasoning text. \\[0.5em]

\texttt{STEP 1: <step text> | <BLOOM\_LEVEL> | BLOOM\_REASON: <brief explanation>} \\
\texttt{STEP 2: <step text> | <BLOOM\_LEVEL> | BLOOM\_REASON: <brief explanation>} \\
\hline
\end{tabular}
\caption{Prompt template used for Annotation (used for both mathematical datasets and BBH tasks). Few-shot examples are omitted due to space constraints, please see the corresponding tables.}
\label{tab:annotation_prompt}
\end{table*}

\begin{table*}[t]
\centering
\small
\begin{tabular}{p{0.95\linewidth}}
\hline
\textbf{Annotation Prompt Few-shot examples (Mathematical Datasets)} \\ 
\hline
Example 1 \\
Question: \\
Julie is reading a 120-page book. Yesterday, she was able to read 12 pages and today, she read twice as many pages as yesterday. If she wants to read half of the remaining pages tomorrow, how many pages should she read?\\

Reasoning trace:\\
Okay, let's tackle this problem step by step. First, I need to figure out how many pages Julie has read so far and then determine how many are left. The book is 120 pages total. Yesterday, she read 12 pages. Today, she read twice as many as yesterday. So, today's pages would be 2 times 12, which is 24 pages. Let me add those up: 12 + 24 = 36 pages read in total over the two days. Now, subtract that from the total pages to find the remaining pages. 120 minus 36 equals 84 pages left. The question says she wants to read half of the remaining pages tomorrow. So, half of 84 is 42. Therefore, she should read 42 pages tomorrow. Let me double-check to make sure I didn't make a mistake. Total pages: 120. Yesterday: 12. Today: 24. Total read: 36. Remaining: 84. Half of 84 is 42. Yep, that seems right.\\

Correct segmentation:\\
STEP 1: Okay, let's tackle this problem step by step. First, I need to figure out how many pages Julie has read so far and then determine how many are left. | ANALYZE | BLOOM\_REASON: The solver breaks the overall task into subproblems (finding pages read so far and pages remaining) and identifies how these parts relate to the goal of solving the problem.\\
STEP 2: The book is 120 pages total. Yesterday, she read 12 pages. Today, she read twice as many as yesterday. | REMEMBER | BLOOM\_REASON: The solver restates numerical facts given in the problem (total pages, pages read yesterday, and the relationship for today) without any transformation or interpretation.\\
STEP 3: So, today's pages would be 2 times 12, which is 24 pages. Let me add those up: 12 + 24 = 36 pages read in total over the two days. Now, subtract that from the total pages to find the remaining pages. 120 minus 36 equals 84 pages left. | APPLY | BLOOM\_REASON: The solver executes a sequence of arithmetic procedures (multiplication, addition, and subtraction) to compute pages read so far and the number of remaining pages.\\
STEP 4: The question says she wants to read half of the remaining pages tomorrow. | REMEMBER | BLOOM\_REASON: The solver restates a condition directly from the problem (reading half of the remaining pages) without performing any operation on it. \\
STEP 5: So, half of 84 is 42. Therefore, she should read 42 pages tomorrow. | APPLY | BLOOM\_REASON: The solver applies a procedure by calculating half of the remaining pages to determine the required number of pages to read.\\
STEP 6: Let me double-check to make sure I didn't make a mistake. Total pages: 120. Yesterday: 12. Today: 24. Total read: 36. Remaining: 84. Half of 84 is 42. Yep, that seems right. | EVALUATE | BLOOM\_REASON: The solver checks the correctness of the solution by verifying intermediate results and confirming that the final answer is consistent with the problem conditions.\\
Example 2\\
Question:\\
Michael had 58 golf balls. On Tuesday, he lost 23 golf balls. On Wednesday, he lost 2 more. How many golf balls did he have at the end of Wednesday? \\

Reasoning trace: \\
Okay, let's tackle this problem step by step. So, Michael starts with 58 golf balls. On Tuesday, he lost 23. Then on Wednesday, he lost 2 more. I need to find out how many he has left by the end of Wednesday. First, let's figure out how many he had after Tuesday. He started with 58 and lost 23. So, subtract 23 from 58. Let me do that calculation. 58 minus 20 is 38, and then minus 3 more would be 35. Wait, 58 - 23. Let me check again: 58 minus 20 is 38, then subtract 3 more (since 23 is 20+3), so 38 - 3 = 35. Yeah, that's right. So after Tuesday, he has 35 golf balls left. Then on Wednesday, he lost 2 more. So now, take the 35 and subtract 2. 35 - 2 = 33. So, after Wednesday, he should have 33 golf balls. Let me double-check to make sure I didn't make a mistake. Starting with 58, lose 23: 58 - 23. Let's do it another way. 23 + 35 = 58, so yes, 58 -23 is 35. Then 35 -2 is 33. That seems right. So the answer should be 33. \\

Correct segmentation: \\
STEP 1: Okay, let's tackle this problem step by step. So, Michael starts with 58 golf balls. On Tuesday, he lost 23. Then on Wednesday, he lost 2 more. | REMEMBER | BLOOM\_REASON: The solver restates facts given in the problem without any transformation or interpretation. \\    
STEP 2: I need to find out how many he has left by the end of Wednesday. | UNDERSTAND | BLOOM\_REASON: The solver interprets the goal of the problem by expressing what needs to be determined in their own words. \\
STEP 3: First, let's figure out how many he had after Tuesday. | ANALYZE | BLOOM\_REASON: The solver breaks the problem into parts by identifying an intermediate quantity (the number of golf balls after Tuesday) needed to reach the final answer. \\
STEP 4: He started with 58 and lost 23. So, subtract 23 from 58. Let me do that calculation. 58 minus 20 is 38, and then minus 3 more would be 35. | APPLY | BLOOM\_REASON: The solver executes arithmetic procedures by performing subtraction to compute the number of golf balls remaining after Tuesday. \\
STEP 5: Wait, 58 - 23. Let me check again: 58 minus 20 is 38, then subtract 3 more (since 23 is 20+3), so 38 - 3 = 35. Yeah, that's right. So after Tuesday, he has 35 golf balls left. | EVALUATE | BLOOM\_REASON: The solver checks the correctness of the calculation by recomputing the subtraction and confirming that the result is accurate. \\
STEP 6: Then on Wednesday, he lost 2 more. So now, take the 35 and subtract 2. 35 - 2 = 33. So, after Wednesday, he should have 33 golf balls. | APPLY | BLOOM\_REASON: The solver executes arithmetic procedures by subtracting the additional loss from the intermediate result to determine the number of golf balls remaining after Wednesday. \\
STEP 7: Let me double-check to make sure I didn't make a mistake. Starting with 58, lose 23: 58 - 23. Let's do it another way. 23 + 35 = 58, so yes, 58 -23 is 35. Then 35 -2 is 33. That seems right. So the answer should be 33. | EVALUATE | BLOOM\_REASON: The solver verifies the correctness of the solution by rechecking calculations using an alternative method and confirming that the final answer is consistent. \\

\hline
\end{tabular}
\caption{Few-shot examples used in the Annotation Prompt for Mathematical datasets}
\label{tab:annotation_few_shpt_examples}
\end{table*}

\begin{table*}[t]
\centering
\small
\begin{tabular}{p{0.95\linewidth}}
\hline
\textbf{CoT Prompt Template (BBH Tasks)} \\ 
\hline
Answer the following question step by step.
End your response by stating the chosen option in the format: "The answer is X.", where X is the correct option.  \\[4pt]

Question: \texttt{\{question\}} \\
\hline
\end{tabular}
\caption{Prompt template used for Chain of Thought generation (BBH Tasks)}
\label{tab:cot_prompt_bbh}
\end{table*}

\begin{table*}[t]
\centering
\small
\begin{tabular}{p{0.95\linewidth}}
\hline
\textbf{Annotation Prompt One-shot example (BBH Formal Fallacies)} \\ 
\hline
Example \\
Question:\\
It is not always easy to grasp who is consuming which products. The following argument pertains to this question: Every infrequent user of Paul Mitchell shampoo is either a rare consumer of Nioxin shampoo or a loyal buyer of Caress soap, or both. No regular consumer of Lush soap is a rare consumer of Nioxin shampoo and, in the same time, a loyal buyer of Caress soap. It follows that whoever is an infrequent user of Paul Mitchell shampoo is not a regular consumer of Lush soap." Is the argument, given the explicitly stated premises, deductively valid or invalid? Options: - valid - invalid \\

Reasoning trace: [omitted due to space constraints] \\
Correct segmentation: \\
STEP 1:  Okay, let's try to figure out if this argument is deductively valid. So, the question is about whether the conclusion follows logically from the premises. Let me break it down step by step. | REMEMBER | BLOOM\_REASON: The solver states what's asked in the question. \\
STEP 2: First, let's parse the premises. The first premise says: "Every infrequent user of Paul Mitchell shampoo is either a rare consumer of Nioxin shampoo or a loyal buyer of Caress soap, or both." So, if someone is an infrequent user of Paul Mitchell (let's call this P), then they are either a rare consumer of Nioxin (N) or a loyal buyer of Caress (C), or both. In logical terms, that's $P \rightarrow (N \lor C)$. | ANALYZE | BLOOM\_REASON: The solver breaks down the first premise into parts and determines how they relate to each other in logic terms.\\
STEP 3: The second premise: "No regular consumer of Lush soap is a rare consumer of Nioxin shampoo and, in the same time, a loyal buyer of Caress soap." So, if someone is a regular consumer of Lush soap (L), then they cannot be both a rare consumer of Nioxin (N) and a loyal buyer of Caress (C). In logical terms, this would be $L \rightarrow \neg (N \land C)$. | ANALYZE | BLOOM\_REASON: The solver breaks down the second premise into parts and determines how they relate to each other in logic terms.\\
STEP 4: Which is equivalent to $L \rightarrow (\neg N \lor \neg C)$ by De Morgan's law. | APPLY | BLOOM\_REASON: The solver applies De Morgan's Law. \\
STEP 5: The conclusion is: "Whoever is an infrequent user of Paul Mitchell shampoo is not a regular consumer of Lush soap." So, $P \rightarrow \neg L$. | UNDERSTAND | BLOOM\_REASON: The solver interprets the conclusion in logical terms. \\
STEP 6: We need to check if from $P \rightarrow (N \lor C)$ and $L \rightarrow \neg (N \land C)$, we can derive $P \rightarrow \neg L$. | UNDERSTAND | BLOOM\_REASON: The solver interprets what should be done. \\
STEP 7: Hmm. Let's see. Let's assume P is true. Then from the first premise, $N \lor C$ is true. Now, suppose for contradiction that L is true. Then from the second premise, $\neg (N \land C)$ must be true, so either $\neg N \lor \neg C$. But since $N \lor C$ is true (from P), if both N and C were false, then $N \lor C$ would be false. Wait, but if $N \lor C$ is true, then at least one of N or C is true. But if L is true, then $\neg (N \land C)$, so they can't both be true. So if $N \lor C$ is true and $\neg (N \land C)$ is true, that means exactly one of N or C is true (exclusive OR). But how does that relate to L? Wait, the conclusion is that P implies not L. So if P is true, then L must be false. Let's see if the premises lead to that. | ANALYZE | BLOOM\_REASON: The solver examines how the premises' logical parts interact and decomposes the relationship between $N \lor C$ and $\neg (N \land C)$. \\
STEP 8: Suppose someone is a P (infrequent user of Paul Mitchell). Then they are $N \lor C$. Now, if they were also an L (regular consumer of Lush), then from the second premise, they can't be $N \land C$. But since they are $N \lor C$, being L would require that they are not both N and C. But $N \lor C$ is true, so if they are L, then they must be exactly one of N or C. | APPLY | BLOOM\_REASON: The solver applies logical rules. \\
STEP 9: But does that lead to a contradiction? Wait, the conclusion is that P implies $\neg L$. So we need to see if P leads to $\neg L$. Let's assume P and L, and see if that leads to a contradiction. Assume $P \land L$. Then from first premise, $N \lor C$. From second premise, since L is true, $\neg (N \land C)$. So we have $N \lor C$ and $\neg (N \land C)$. That's possible; for example, N is true and C is false, or vice versa. So there's no contradiction here. Therefore, it's possible for someone to be P, L, and have either N or C but not both. So the assumption of $P \land L$ doesn't lead to a contradiction. Therefore, the conclusion $P \rightarrow \neg L$ doesn't necessarily follow. | EVALUATE | BLOOM\_REASON: The solver checks whether there is a contradiction. \\
STEP 10: Wait, but the argument says 'it follows that whoever is an infrequent user of Paul Mitchell shampoo is not a regular consumer of Lush soap'. So if there exists someone who is P and L, then the conclusion is false.  | UNDERSTAND | BLOOM\_REASON: The solver interprets a part of the question. \\
STEP 11: But according to the premises, can such a person exist? Let's construct a scenario. Let's say someone is an infrequent user of Paul Mitchell (P), so they are either N or C (or both). Suppose they are a rare consumer of Nioxin (N) but not a loyal buyer of Caress (so C is false). Then $N \lor C$ is true (since N is true). Now, they are a regular consumer of Lush (L). The second premise says that L implies $\neg (N \land C)$. Since N is true and C is false, $N \land C$ is false, so $\neg (N \land C)$ is true. So L can be true here. So this person is P, L, N, $\neg C$. So they satisfy both premises but the conclusion ($P \rightarrow \neg L$)) would be false for them because they are P and L. Therefore, the conclusion doesn't hold in this case. So the argument is invalid. | CREATE | BLOOM\_REASON: The solver constructs an original counterexample with specific truth-value assignments to demonstrate invalidity. \\
STEP 12: Wait, but the first premise says "every infrequent user of Paul Mitchell is either a rare consumer of Nioxin or a loyal buyer of Caress or both." So in this example, the person is a rare consumer of Nioxin (so N is true), so they satisfy the first premise. The second premise says no regular consumer of Lush is both N and C. In this case, the person is L, N, and $\neg C$, so they are not both N and C, so the second premise is satisfied. Therefore, the premises don't prevent someone from being P and L. Hence, the conclusion that P implies $\neg L$ is not necessarily true. Therefore, the argument is invalid. So the answer should be invalid.| EVALUATE | BLOOM\_REASON: The solver verifies the counterexample against each premise and judges the argument as invalid.\\
\hline
\end{tabular}
\caption{One-shot example used in the Annotation Prompt for BBH Formal Fallacies}
\label{tab:annotation_few_shpt_example_bbh}
\end{table*}

\begin{table*}[t]
\centering
\small
\begin{tabular}{p{0.95\linewidth}}
\hline
\textbf{Annotation Prompt One-shot example (BBH Hyperbaton)} \\ 
\hline
Example \\
Question:\\
Which sentence has the correct adjective order: Options:(A) lovely tiny modern French restaurant (B) French modern tiny lovely restaurant \\

Reasoning trace: [omitted due to space constraints] \\
Correct segmentation: \\
STEP 1: Okay, the user is asking about correct adjective order in English. They've given two options for describing a restaurant and want to know which one follows the standard adjective sequence rules. | UNDERSTAND | BLOOM\_REASON: The solver interprets the task and identifies that the question concerns adjective ordering in English grammar. \\
STEP 2: Hmm, this seems like a grammar question testing knowledge of the ''royal order'' of adjectives. | UNDERSTAND | BLOOM\_REASON: The solver classifies the type of problem and connects it to a known grammatical concept. \\
STEP 3: I recall that in English, adjectives follow a specific hierarchy: opinion, size, age, shape, color, origin, material, purpose. | REMEMBER | BLOOM\_REASON: The solver recalls the standard adjective order rule from prior knowledge. \\
STEP 4: Let me mentally run through the options. Option A: ''lovely tiny modern French restaurant'' — ''lovely'' = opinion (good), ''tiny'' = size (small), ''modern'' = age (new), ''French'' = origin, ''restaurant'' = noun. | ANALYZE | BLOOM\_REASON: The solver decomposes Option A into grammatical categories and examines how its parts relate to the adjective hierarchy. \\
STEP 5: Option B: ''French modern tiny lovely restaurant'' — ''French'' = origin (should come later), ''modern'' = age (should come before origin), ''tiny'' = size (should come before age), ''lovely'' = opinion (should come first), ''restaurant'' = noun. | ANALYZE | BLOOM\_REASON: The solver breaks down Option B and compares the adjective positions against the expected ordering. \\
STEP 6: Yes, the standard order is opinion $\rightarrow$ size $\rightarrow$ age $\rightarrow$ origin. | REMEMBER | BLOOM\_REASON: The solver restates the grammatical rule needed for verification. \\
STEP 7: So ''lovely'' (opinion) should come first, then ''tiny'' (size), then ''modern'' (age), then ''French'' (origin). | APPLY | BLOOM\_REASON: The solver applies the adjective-ordering rule to determine the correct sequence. \\
STEP 8: That matches option A perfectly. | EVALUATE | BLOOM\_REASON: The solver judges that Option A satisfies the grammatical criteria. \\
STEP 9: Option B has origin (''French'') at the very beginning, which is wrong because origin adjectives usually come later. Also, ''modern'' (age) is misplaced before ''tiny'' (size), which should come first. | EVALUATE | BLOOM\_REASON: The solver critiques Option B by checking it against the grammatical rule and identifying violations. \\
STEP 10: The user seems to be learning English grammar, probably a student or non-native speaker. They're being careful by asking for step-by-step reasoning, which is good. I should make sure my explanation is clear but not overwhelming. | UNDERSTAND | BLOOM\_REASON: The solver interprets the user’s likely needs and adjusts the communication approach accordingly. \\
STEP 11: Yes, I'm confident. The correct order is A. The answer should be A. I'll state it plainly at the end as requested. | EVALUATE | BLOOM\_REASON: The solver confirms the correctness of the conclusion before presenting the final answer. \\
\\
\hline
\end{tabular}
\caption{One-shot example used in the Annotation Prompt for BBH Hyperbaton}
\label{tab:annotation_few_shpt_example_bbh_hyperbaton}
\end{table*}

\section{A closer look at the outlier Phi-4-Reasoning}
\label{sec:phi-analysis}
\begin{figure*}[t]
\centering
\includegraphics[width=\textwidth]{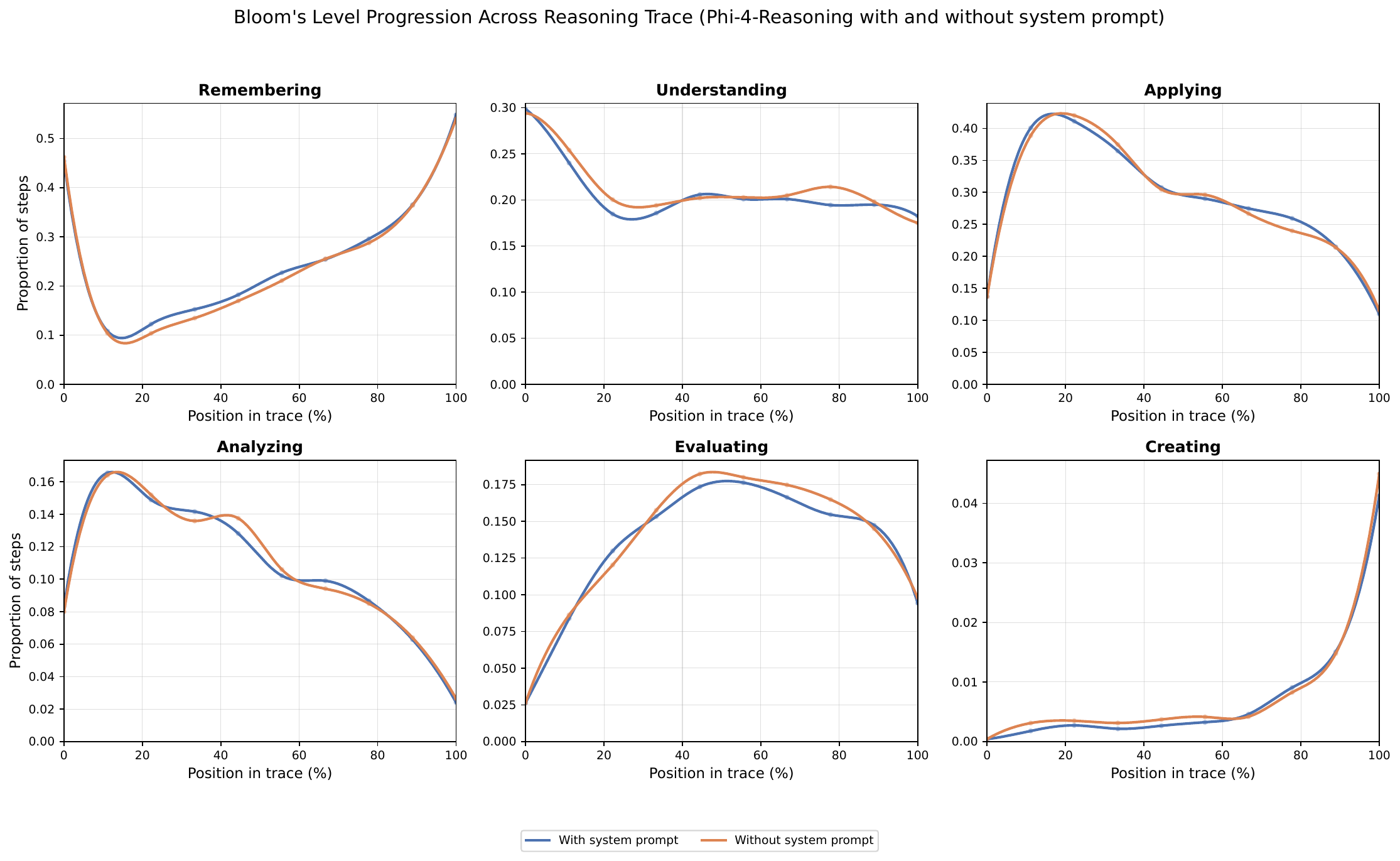}
\caption{Temporal Dynamics of Bloom's Levels in Reasoning Traces (Phi-4-reasoning is ran with and without system prompt- internal reasoning traces only)}
\label{fig:temporal_dynamics_phi_no_system_vs_phi}
\vspace{-1em}
\end{figure*}

Phi-4-Reasoning’s divergent thinking profile warrants further analysis, in which we emphasize two aspects: first, we examine whether this behavior is enforced by the system prompt provided by the model developers; second, we conduct a small-scale qualitative analysis of selected examples to gain deeper insight into this phenomenon. As far as the first aspect is concerned, we reran the experiments without the system prompt, and the resulting distribution and temporal dynamics can be found in Table~\ref{tab:phi4_system_prompt} and fig. ~\ref{fig:temporal_dynamics_phi_no_system_vs_phi}. Removing the system prompt only marginally shifts the distribution (e.g., Remembering 28.4\% → 27.6\%, Evaluating 12.7\% → 13.0\%), while the overall profile and temporal dynamics are preserved: Remembering remains by far the highest across models with a rise at the end, Applying the lowest, and Creating nonzero. The outlier cognitive profile is thus a property of the model rather than an artifact of the system prompt. While we cannot pinpoint the exact cause, several aspects of Phi-4-Reasoning's training \citep{abdin2025phi4reasoningtechnicalreport} offer plausible explanations: a different teacher model (o3-mini), SFT-only training without an outcome-based RL stage, and a heavily curated, knowledge-dense training corpus. A small-scale qualitative inspection of traces, focused on the most striking divergence---Remembering at the end of the trace---reveals that (with both prompt settings) the model frequently terminates by repeatedly restating its intended final answer and output format (e.g., "I'll produce answer with final line: \"The answer is 540.", "We'll produce a final answer message with that format: \"The answer is X."). This terminal-repetition pattern is consistent with the absence of an outcome-based RL stage, which would otherwise pressure clean termination.
\begin{table*}[t]
\centering
\begin{tabular}{lcccccc}
\toprule
\textbf{Setting} & \textbf{Remember} & \textbf{Understand} & \textbf{Apply} & \textbf{Analyze} & \textbf{Evaluate} & \textbf{Create} \\
\midrule
With system prompt    & 28.4 & 21.0 & 26.9 & 10.1 & 12.7 & 0.9  \\
Without system prompt & 27.6 & 21.5 & 26.8 & 10.1 & 13.0 & 1.0\\
\bottomrule
\end{tabular}
\caption{Distribution comparison for Phi-4-Reasoning with and without the default system prompt (internal reasoning traces only).}
\label{tab:phi4_system_prompt}
\end{table*}

\section{Example Outputs across Diverse Tasks}
\label{sec:example_outputs}
In this section, we present example outputs of our annotation framework across diverse tasks in order to demonstrate the differences in characteristics across tasks (Tables ~\ref{tab:example_math}, ~\ref{tab:example_fallacies}, ~\ref{tab:example_hyperbaton}).

\begin{table*}[t]
\centering
\small

\begin{tabularx}{\textwidth}{|X|p{2.5cm}|X|}
\hline

\multicolumn{3}{|p{\dimexpr\textwidth-2\tabcolsep-2\arrayrulewidth}|}{
\textbf{Question:}

Mike plays ping pong for 40 minutes. In the first 20 minutes, he scores 4 points. In the second 20 minutes, he scores 25\% more points. How many total points did he score?
}
\\
\hline

\textbf{Reasoning Text} &
\textbf{Label} &
\textbf{Justification}
\\
\hline

Okay, let's see. Mike plays ping pong for 40 minutes. The problem says in the first 20 minutes, he scores 4 points. Then in the second 20 minutes, he scores 25\% more points. I need to find the total points.
&
REMEMBERING
&
The solver restates facts given in the problem without any transformation or interpretation.
\\
\hline

First, let's confirm the time: 40 minutes total, split into two 20-minute intervals. First 20 minutes: 4 points. Second 20 minutes: 25\% more than the first 20 minutes.
&
UNDERSTANDING
&
The solver interprets the problem by expressing the given information in their own words and identifying the key elements.
\\
\hline

Wait, does ``25\% more points'' mean 25\% more than the first 20 minutes' points? I think that's what it means. So first part is 4 points, second part is 4 plus 25\% of 4.
&
UNDERSTANDING
&
The solver clarifies the meaning of ``25\% more points'' and relates it to the first 20 minutes' points.
\\
\hline

Let me calculate that. 25\% of 4 is $0.25 \times 4 = 1$. So 25\% more would be $4 + 1 = 5$ points in the second 20 minutes.
&
APPLYING
&
The solver executes a calculation to find 25\% of 4 and then applies this result to determine the points scored in the second 20 minutes.
\\
\hline

Then total points would be $4 + 5 = 9$.
&
APPLYING
&
The solver applies basic arithmetic to calculate the total points scored.
\\
\hline

Wait, let me make sure. 25\% more than 4. So $4 \times 1.25 = 5$. Yeah, that's right. So second 20 minutes is 5 points. Total is $4 + 5 = 9$.
&
EVALUATING
&
The solver checks the calculation by using an alternative method (multiplication by 1.25) and confirms that the result is accurate.
\\
\hline

Let me double-check. First 20 minutes: 4 points. Second 20 minutes: 25\% more than first. 25\% of 4 is 1, so $4 + 1 = 5$. Total $4 + 5 = 9$. That seems right. I don't think there's any trick here. The problem says ``in the second 20 minutes, he scores 25\% more points'' -- more than the first 20 minutes, which is 4. So yes, 25\% more than 4 is 5. Total points 9.
&
EVALUATING
&
The solver verifies the correctness of the solution by rechecking calculations and confirming that the final answer is consistent with the problem conditions.
\\
\hline

\end{tabularx}

\caption{Example of an annotated sample (GSM8K)}
\label{tab:example_math}

\end{table*}

\begin{table*}[t]
\centering
\small

\begin{tabularx}{\textwidth}{|X|p{2.5cm}|X|}
\hline

\hline
\multicolumn{3}{|p{\dimexpr\textwidth-2\tabcolsep-2\arrayrulewidth}|}{
\textbf{Question:}

``Here comes a perfectly valid argument: First, being a cousin of Chris is sufficient for not being a son of Kermit. We may conclude that whoever is not a son of Kermit is a cousin of Chris.''

Is the argument, given the explicitly stated premises, deductively valid or invalid? Options: \textit{valid} / \textit{invalid}.
}
\\
\hline

Okay, let's try to figure out if this argument is deductively valid. So, the question says: ``First, being a cousin of Chris is sufficient for not being a son of Kermit. We may conclude that whoever is not a son of Kermit is a cousin of Chris.'' &
REMEMBER &
The solver recalls the question and the argument to be evaluated.
\\
\hline

Hmm, let's break this down. The first part is a premise: ``being a cousin of Chris is sufficient for not being a son of Kermit.'' In logical terms, if someone is a cousin of Chris (let's call that $C$), then they are not a son of Kermit (not $S$). So, $C \rightarrow \neg S$.&
ANALYZE &
The solver breaks down the premise into logical terms and identifies the relationship between being a cousin of Chris and not being a son of Kermit.
\\
\hline

The conclusion is: ``whoever is not a son of Kermit is a cousin of Chris.'' So, $\neg S \rightarrow C$. &
UNDERSTAND &
The solver interprets the conclusion in logical terms and identifies the relationship between not being a son of Kermit and being a cousin of Chris.
\\
\hline

Wait a second, the premise is $C \rightarrow \neg S$, but the conclusion is $\neg S \rightarrow C$. That looks like the converse of the premise. In logic, the converse of an implication isn't necessarily true. For example, if ``if it's a dog, then it's a mammal'' ($D \rightarrow M$), the converse would be ``if it's a mammal, then it's a dog'' ($M \rightarrow D$), which isn't true because there are mammals that aren't dogs. &
ANALYZE &
The solver recognizes the relationship between the premise and the conclusion, and identifies the converse of the premise.
\\
\hline

So here, the premise is $C$ implies $\neg S$. The conclusion is saying $\neg S$ implies $C$. That's the converse, which isn't valid. So the argument is assuming that because $C$ leads to $\neg S$, then $\neg S$ must lead to $C$. But that's a fallacy. It's like saying ``if it's raining, the ground is wet'' ($R \rightarrow W$), so ``if the ground is wet, it's raining'' ($W \rightarrow R$), which isn't necessarily true because the ground could be wet for other reasons. & 
EVALUATE &
The solver evaluates the argument and identifies the fallacy of assuming the converse is true.
\\
\hline

Let me check again. The premise: $C \rightarrow \neg S$. The conclusion: $\neg S \rightarrow C$. So the argument is taking $C \rightarrow \neg S$ and concluding $\neg S \rightarrow C$. But that's not a valid deduction. The correct contrapositive of $C \rightarrow \neg S$ would be $S \rightarrow \neg C$, right? Because the contrapositive of $P \rightarrow Q$ is $\neg Q \rightarrow \neg P$. So here, the contrapositive would be $S \rightarrow \neg C$. But the conclusion is $\neg S \rightarrow C$, which is different. &
ANALYZE &
The solver re-examines the premise and conclusion, and identifies the correct contrapositive.
\\
\hline

So the argument is confusing the implication with its converse. Therefore, the argument is invalid because the conclusion doesn't logically follow from the premise. The premise says being a cousin of Chris ensures you're not a son of Kermit, but it doesn't say that everyone who isn't a son of Kermit must be a cousin of Chris. There could be other people who aren't sons of Kermit but aren't cousins of Chris either. &
EVALUATE &
The solver evaluates the argument and concludes that it is invalid.
\\
\hline

For example, suppose Kermit has a daughter who's not a son, so she's not a son of Kermit, but she's not a cousin of Chris. The premise says cousins of Chris aren't sons of Kermit, but it doesn't say that all non-sons are cousins. So the conclusion is making a broader claim than the premise supports. Hence, invalid. &
CREATE &   
The solver constructs a counterexample to demonstrate the invalidity of the argument.

\end{tabularx}

\caption{Example of an annotated sample (BBH formal fallacies)}
\label{tab:example_fallacies}
\end{table*}

\begin{table*}[t]
\centering
\small

\begin{tabularx}{\textwidth}{|X|p{2.5cm}|X|}
\hline

\multicolumn{3}{|p{\dimexpr\textwidth-2\tabcolsep-2\arrayrulewidth}|}{
\textbf{Question:}

\emph{Which sentence has the correct adjective order: Options: (A) silly old prismlike Mexican sock (B) Mexican prismlike old silly sock}
}
\\
\hline

\textbf{Reasoning Text} &
\textbf{Label} &
\textbf{Justification}
\\
\hline

Okay, the user is asking about adjective order in English, specifically which sentence correctly orders adjectives. The options are: (A) ``silly old prismlike Mexican sock''
(B) ``Mexican prismlike old silly sock''. & UNDERSTAND & The solver interprets the task and identifies the question concerns adjective ordering in English grammar.\\
\hline

Hmm, I recall that English has a specific sequence for multiple adjectives before a noun. It's not random—it follows a pattern: opinion, size, age, shape, color, origin, material, purpose. & REMEMBER & The solver recalls the standard adjective order rule from prior knowledge.\\
\hline

Let me mentally run through the rules. First, ``silly'' is an opinion adjective---it describes how someone feels about the sock. Then ``old'' is clearly about age. ``Prismlike'' is a shape descriptor, and ``Mexican'' indicates origin. & ANALYZE & The solver breaks down the adjectives into their respective categories and examines how they relate to the adjective hierarchy. \\
\hline

So the correct order should be opinion $\rightarrow$ age $\rightarrow$ shape $\rightarrow$ origin. That would be ``silly old prismlike Mexican sock.''& APPLY & The solver applies the adjective-ordering rule to determine the correct sequence. \\
\hline

Option (B) has ``Mexican'' first, which is origin, but origin should come after shape, not before. It also has ``prismlike'' before ``old,'' which messes up the age order. ``Old'' should come before ``prismlike'' because age comes before shape in the sequence. & EVALUATE & The solver critiques Option B by checking it against the grammatical rule and identifying violations. \\
\hline

Wait, let me double-check the standard order: Opinion $\rightarrow$ Size $\rightarrow$ Age $\rightarrow$ Shape $\rightarrow$ Color $\rightarrow$ Origin $\rightarrow$ Material $\rightarrow$ Purpose. & REMEMBER & The solver restates the grammatical rule needed for verification.\\
\hline

Yep, so ``silly'' (opinion) $\rightarrow$ ``old'' (age) $\rightarrow$ ``prismlike'' (shape) $\rightarrow$ ``Mexican'' (origin). Option A matches this perfectly. Option B has origin first, then shape, then age, then opinion---total chaos. & EVALUATE & The solver judges that Option A satisfies the grammatical criteria and further critiques Option B. \\
\hline

The user seems to be learning English grammar, probably a student or non-native speaker. They're being careful by asking for step-by-step reasoning, which is smart. I should make sure my explanation is clear but not overwhelming. & UNDERSTAND & The solver interprets the user’s likely needs and adjusts the communication approach accordingly.\\
\hline

Also, the sock example is oddly specific---prismlike? That's a fun detail. Maybe it's a typo for ``prism-like'' or a creative descriptor, but the grammar still holds. & UNDERSTAND & The solver notes an unusual aspect of the example but recognizes it does not affect the grammatical analysis. \\
\hline

Final check: Option A = silly (opinion) + old (age) + prismlike (shape) + Mexican (origin). Correct. Option B = Mexican (origin) + prismlike (shape) + old (age) + silly (opinion). Wrong order. & EVALUATE & The solver conducts a final verification of the options against the grammatical rule. \\
\hline

Yep, A is definitely right. I’ll wrap up by stating it plainly at the end as requested. & EVALUATE & The solver confirms the correctness of the conclusion before presenting the final answer.\\
\hline

\end{tabularx}

\caption{Example of an annotated sample (BBH hyperbaton)}
\label{tab:example_hyperbaton}

\end{table*}

\section{Human Annotation Details}
\label{sec:human_annotation_details}
\begin{figure*}[t]
\centering
\includegraphics[width=\textwidth]{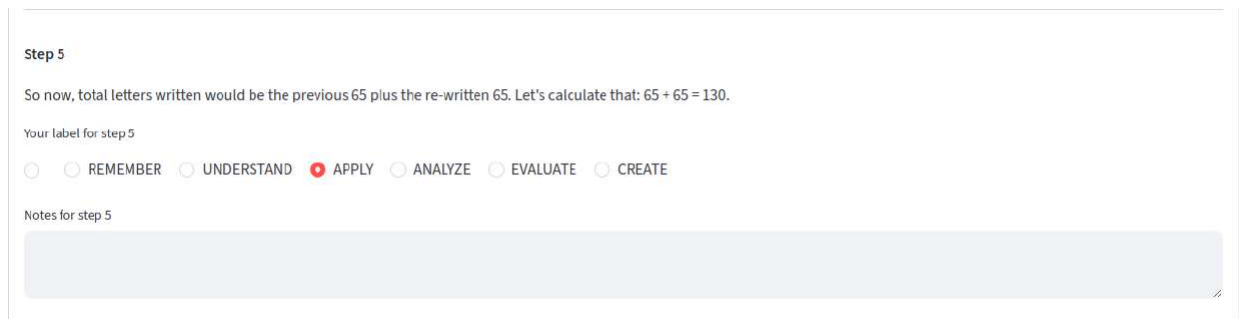}
\caption{Human Annotation UI}
\label{fig:human_annotation_ui}
\vspace{-1em}
\end{figure*}

In this section, we describe the annotation process in detail. First, the authors introduced the volunteer non-author annotator to the project's objectives and task requirements. Although the annotator was already familiar with Bloom’s taxonomy, we also provided general background information on the topic (similar to the material presented in Section 3 of the paper). Finally, both annotators—the author and the non-author annotator—completed the annotations independently using the same prompt that was provided to the LLM annotator through a user interface designed to facilitate the annotation process (see fig ~\ref{fig:human_annotation_ui}).

\end{document}